\documentclass[letterpaper,journal]{IEEEtran}
\usepackage{amsmath,amsfonts}
\usepackage{algorithmic}
\usepackage{algorithm}
\usepackage{array}
\usepackage[caption=false,font=normalsize,labelfont=sf,textfont=sf]{subfig}
\usepackage{textcomp}
\usepackage{stfloats}
\usepackage{url}
\usepackage{verbatim}
\usepackage{graphicx}
\usepackage{cite}
\usepackage{booktabs}
\usepackage{threeparttable}
\usepackage{tabularx}
\usepackage{amssymb}
\usepackage{makecell}
\usepackage{booktabs,multirow,threeparttable,array}
\newcolumntype{M}[1]{>{\centering\arraybackslash}m{#1}}
\usepackage[colorlinks=true,linkcolor=darkred,citecolor=blue,]{hyperref}

\usepackage[english]{babel}
\usepackage{blindtext}
\usepackage[table]{xcolor}
\definecolor{darkred}{RGB}{139,0,0}

\begin{document}

\title{Collaborative Continuum Robots: A Survey from an Automation-System Perspective}

\author{
Xinyu Li,~\IEEEmembership{Graduate Student Member,~IEEE},
Qian Tang,
Guoxin Yin, Gang Zheng,~\IEEEmembership{Senior Member,~IEEE}, Jessica Burgner-Kahrs,~\IEEEmembership{Senior Member,~IEEE}, Cesare Stefanini,~\IEEEmembership{Member,~IEEE}
and Ke Wu,~\IEEEmembership{Member,~IEEE}

\thanks{This work was supported by the National Natural Science Foundation of China (Grant number 52335006). (\textit{Corresponding author: Qian Tang and Ke Wu.})}
\thanks{Xinyu Li, Qian Tang and Guoxin Yin are with the College of Mechanical and Vehicle Engineering, Mechanical Engineering,
	Chongqing University, Chongqing 400044, China (e-mail: xyli@stu.cqu.edu.cn, tqcqu@cqu.edu.cn, gxyin@stu.cqu.edu.cn).}%
\thanks{Cesare Stefanini and Ke Wu are with the Robotics Department,
	Mohamed bin Zayed University of Artificial Intelligence, Abu Dhabi,  United Arab Emirates
	(e-mail: Cesare.Stefanini@mbzuai.ac.ae, ke.wu@mbzuai.ac.ae).}
\thanks{Gang Zheng is with the Defrost Team, Inria, Centrale Lille, CRIStAL - Centre de Recherche en Informatique Signal et Automatique de Lille - UMR 9189, University of Lille, 59000 Lille, France (e-mail: gang.zheng@inria.fr).}
\thanks{Jessica Burgner-Kahrs is with the Continuum Robotics Laboratory, Department of Mathematical and Computational Sciences, University of Toronto, Mississauga, ON L5L 1C6, Canada (e-mail: jessica.burgnerkahrs@utoronto.ca).}

}


\maketitle

\begin{abstract}
Continuum robots (CRs), owing to their compact structure, inherent compliance, and flexible deformation, have been widely applied in various fields. By coordinating multiple CRs to form collaborative continuum robots (CCRs), task adaptability, flexibility and load capacity can be further improved, thus offering significant advantages. In recent years, interest in this emerging field has grown steadily within the continuum-robotics community. In this survey, a closed-loop task-execution architecture for CCRs is constructed from an automation-system perspective, clarifying the system-level logic that transforms target-task specifications into multi-CR collaborative execution. Based on this architecture, CCRs are classified into three representative collaboration modes, namely separated collaboration, assistance collaboration, and parallel collaboration. Research progress is then systematically summarized and analyzed in terms of key technologies, including multi-CR task allocation, structural design, modeling, motion planning methods, control strategies, and sensing mechanisms. Finally, research opportunities in key technologies and system integration challenges for real-world deployment are discussed, providing guidance for research and application of CCRs.
\end{abstract}

\def\abstractname{Note to Practitioners}

\begin{abstract}
Continuum robots are increasingly considered for automation tasks that require access to confined, curved, or fragile environments, yet single-arm systems remain limited in payload capacity, robustness, and task versatility. Collaborative continuum robots (CCRs) address these limitations by coordinating multiple continuum arms through separation, assistance, or parallel coupling. This survey organizes existing CCR research from an automation-system perspective, covering structural design choices, modeling assumptions, motion planning methods, and control strategies relevant to real-world deployment. Practitioners can use this review to identify which collaboration mode best matches application requirements such as load sharing, stiffness modulation, redundancy, and workspace extension, as well as to understand the associated trade-offs in modeling complexity, sensing, and control. The surveyed methods are applicable to automation scenarios including industrial inspection and maintenance, surgical and rehabilitation systems, field robotics, and other environments where adaptability and compliance are critical.
\end{abstract}


\section{INTRODUCTION}
\IEEEPARstart{C}{ontinuum} Robots (CRs), inspired by flexible structures in nature, are characterized by an actuation system decoupled from the deformable body, while maintaining tangent continuity along the backbone during deformation \cite{156,157}. This design provides CRs with a compact structure, inherent compliance, and flexible deformation, enabling effective operation in regions inaccessible to rigid robots \cite{159}. Consequently, CRs have found widespread applications in minimally invasive surgery \cite{158}, equipment repair \cite{160}, and underwater operations \cite{108}.

Despite the increasing deployment of CRs in practical applications, several critical limitations remain. Tasks such as knotting \cite{72}, gripping \cite{155}, and locomotion \cite{142} often exceed the capability of a single CR, constraining task adaptability. Moreover, their high compliance makes them prone to deformation under external loads \cite{119}. Therefore, it is imperative to extend the traditional design paradigm of CRs by enabling multiple CRs to collaborate, thereby forming multi-arm collaborative continuum robots (CCRs) that offer new opportunities for automated execution complex tasks.
\IEEEpubidadjcol

\begin{figure} [t] 
	\centering
	\includegraphics[scale=0.46]{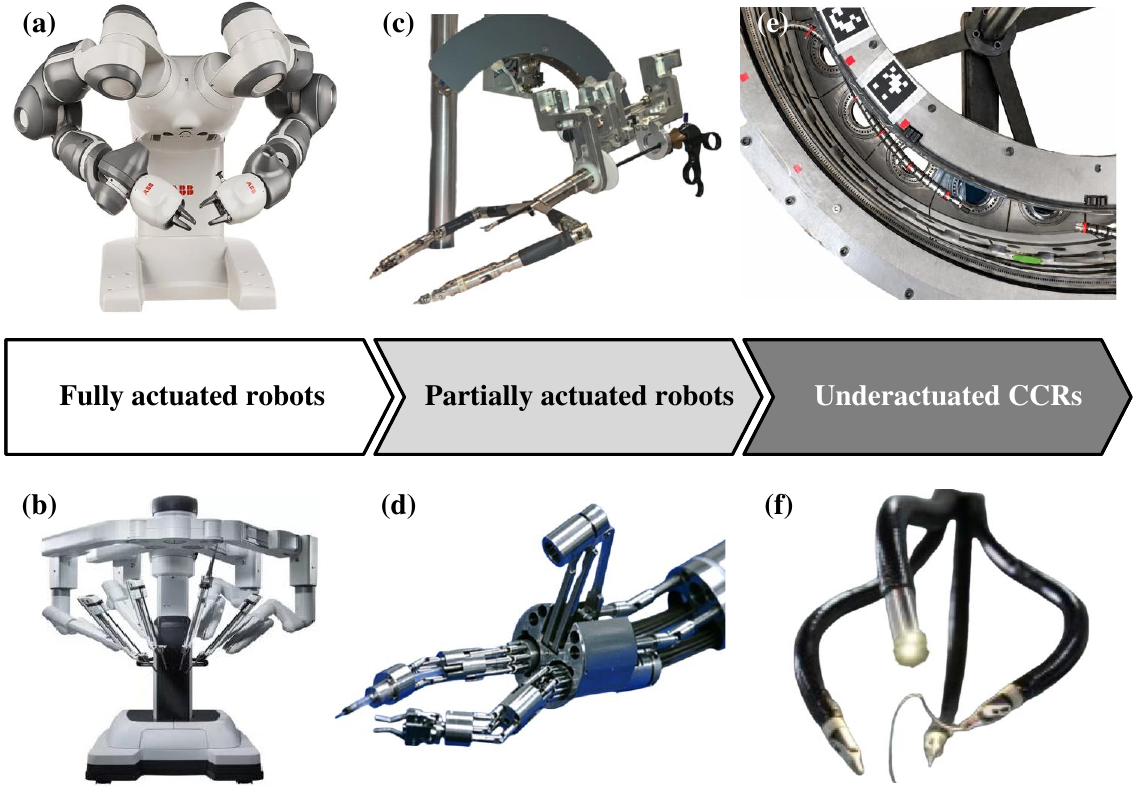} 
	\centering 
	\caption{Development of representative collaborative robots. (a) YuMi dual-arm robot \textcolor{blue}{[13]}. (b) Da Vinci surgical robot \textcolor{blue}{[12]}. (c) The Hong Kong Polytechnic University R\&D surgical robot \textcolor{blue}{[14]}. (d) Waseda University R\&D surgical robot \textcolor{blue}{[15]} . (e) University of Nottingham R\&D repair CCR \textcolor{blue}{[16]}. (f) Shanghai Jiao Tong University R\&D surgical CCR \textcolor{blue}{[17]}.}
	\label{Fig_1} 
\end{figure}

\begin{figure*} [t] 
	\centering
	\includegraphics[scale=0.5]{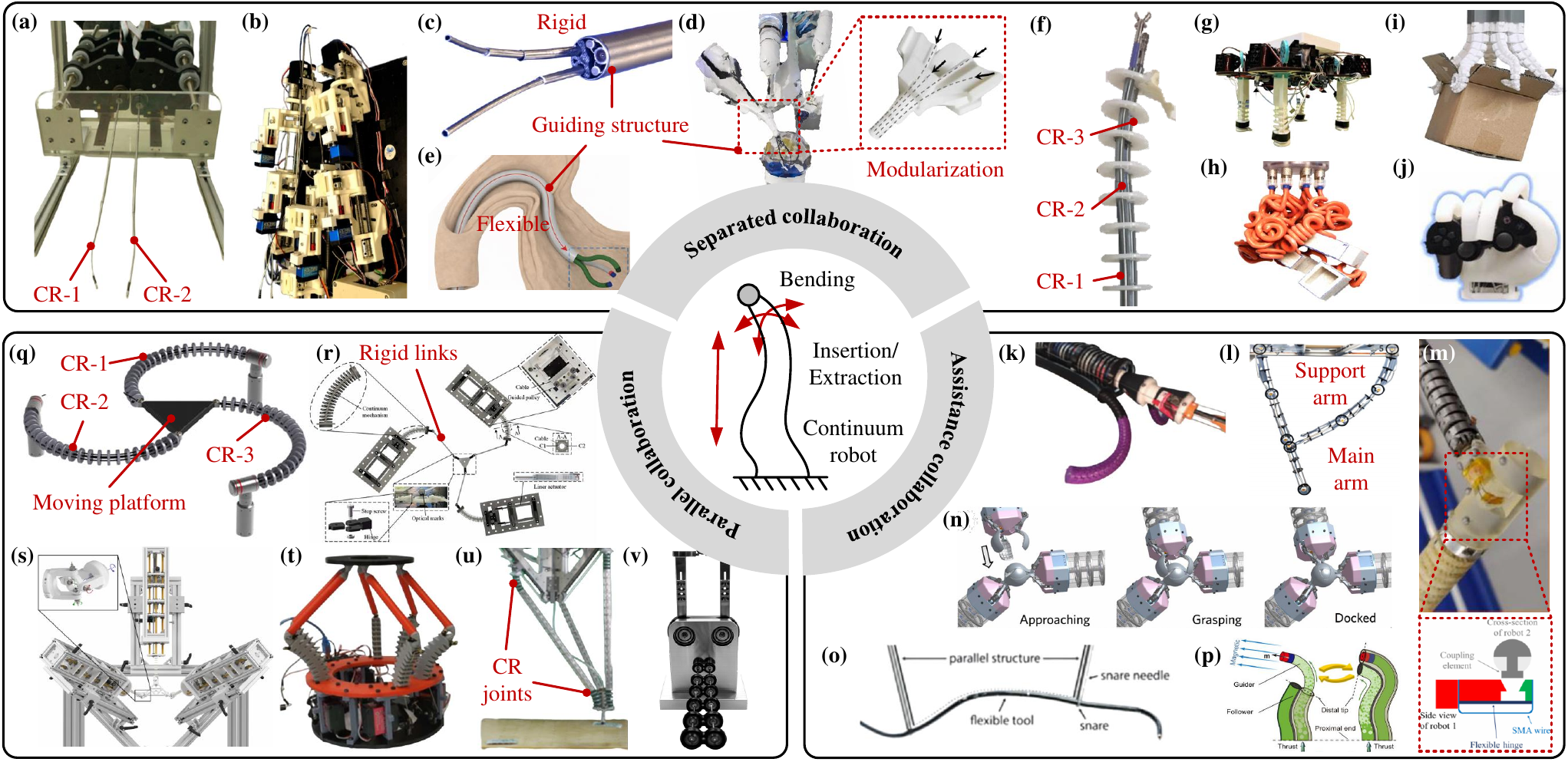} 
	\centering 
	\caption{Classification and representative examples of CCRs. (a) Dual-arm concentric tube robots \cite{18}. (b) Pyramid-type actuation \cite{27}.
	(c) Rigid guiding structures \cite{20}.
		(d) Modular guiding structures \cite{90}.
		(e) Flexible guiding structures \cite{89}.
		(f) Flexible guiding structures \cite{11}.
		(g) CCR-based legged robot \cite{96}. 
		(h) Multiple fluid-driven CRs \cite{111}.
		(i) Logarithmic spiral-based structure \cite{81}. (j) Dexterous hand \cite{103}. (k) Inspired by Angel Oak tree growth \cite{57}.
		(l) Multi-port design \cite{66}.
		(m) Shape-memory alloy connectors \cite{70}.
		(n) Multi-finger spherical gripper \cite{125}.
		(o) Reconfigurable design \cite{60}.
		(p) Alternating follower motion \cite{88}.
		(q) Planar configuration with CRs \cite{78}.
		(r) Planar configuration with CRs and rigid links \cite{73}.
		(s) Spatial configuration with CRs \cite{121}.
		(t) Spatial configuration with high load capacity \cite{119}.
		(u) Composed of CR joints \cite{58}. (v)
		Antagonistic configuration \cite{192}.}
	\label{Fig_2} 
\end{figure*}

The concept of “multi-arm collaboration” was initially introduced in industrial robotics \cite{151}. To address the high-DOF requirements of complex tasks, collaborative robots have progressed from fully actuated rigid systems with few DOFs \cite{152,154}, to partially actuated systems with moderate DOFs \cite{153,131}, and ultimately to underactuated CCRs offering theoretically infinite DOFs \cite{2,90}, as illustrated in Fig.~\ref{Fig_1}. In this process, the collaboration paradigm has evolved from discrete, joint-based coordination to continuous, deformation-driven coupling. Consequently, the coupling relationships among CRs have become a central focus in CCRs research. CCRs are herein classified into three modes, namely separated collaboration, assistance collaboration, and parallel collaboration. Separated collaboration decomposes complex tasks that are difficult for a single CR into simpler subtasks executed cooperatively by multiple CRs, thereby improving adaptability in complex tasks \cite{18} (Figs. \ref{Fig_2}\textcolor{darkred}{(a)}–\ref{Fig_2}\textcolor{darkred}{(j)}). Assistance collaboration physically links CRs to support the master arm, extending its flexibility \cite{60,57} (Figs. \ref{Fig_2}\textcolor{darkred}{(k)}–\ref{Fig_2}\textcolor{darkred}{(p)}). Parallel collaboration attaches the end-effectors of multiple CRs to a common mobile platform, forming a parallel configuration that enhances load capacity while maintaining overall compliance \cite{73} (Fig. \ref{Fig_2}\textcolor{darkred}{(q)}–\ref{Fig_2}\textcolor{darkred}{(v)}). Research on these CCRs is typically based on single-CR frameworks while considering interactions between multiple CRs or between CRs and the environment. 

Most classical reviews in the field of CRs have not treated CCRs as a distinct research topic \cite{195, 196, 206}. Only a limited number of existing reviews (Table~\ref{Tab_CCRSReviews}) have touched upon CCR-related issues in specific contexts. For example, \cite{47} and \cite{197} introduced the structural design of CRs in surgery scenarios, and \cite{41} and \cite{163} further explored their modeling, motion planning and control strategies, some of which can be categorized as separated collaboration in CCR. \cite{76} discussed the structural design and modeling of parallel continuum robots (PCRs), while \cite{52} subsequently classified PCRs into active and passive types, with some active forms considered parallel collaborative. \cite{128}, \cite{164}, \cite{165}, \cite{166}, and \cite{167} reviewed grasper and dexterous-hand structures composed of multiple CRs, while \cite{130} reported on biomimetic soft-legged robots driven by multiple CRs, addressing aspects of structural design, modeling, and control strategies. Several of these designs can also be categorized as separated collaborative. Although a few existing reviews have touched upon CCR-related topics, these discussions remain fragmented and lack a systematic analysis of CCRs as a distinct research topic. Moreover, practical deployment in automated applications has become an important development goal for CCRs \cite{2,6,90}, which requires them to be examined from an automation-system perspective. However, existing reviews have not systematically clarified the collaboration modes, technical organization, and challenges of CCRs within this perspective, making it difficult for readers to understand their research evolution and future development directions. Therefore, a systematic survey of CCRs from an automation-system perspective is urgently needed.

\begin{table*}[t]
	\centering
	\begin{threeparttable}
		\caption{\textbf{Summary of Related Reviews for CCRs}}
		\label{Tab_CCRSReviews}
	\begin{tabularx}{\textwidth}{c c X c c c}
		\toprule
		\textbf{Ref.} & \textbf{Year} & \multicolumn{1}{c}{\textbf{Review Title}} & \textbf{CCR-Related Content} & \textbf{Application} & \textbf{Relevance to CCR} \\
		\midrule
		{\cite{41}} & 2018 & Medical technologies and challenges of robot-assisted minimally invasive intervention and diagnostics & SC: S-M-C & Surgery & H \\
		{\cite{128}} & 2020 & In the soft grip of nature & SC: S & Grasping & L \\
		{\cite{47}} & 2021 & Research progress and development trend of surgical robot and surgical instrument arm & SC: S & Surgery & M \\
		{\cite{197}} & 2022 & Continuum robots for medical interventions & SC: S & Surgery & L \\
		{\cite{163}} & 2022 & Continuum robots for endoscopic sinus surgery: Recent advances, challenges, and prospects & SC: S-P-C & Surgery & H \\
		{\cite{164}} & 2023 & Current designs of robotic arm grippers: A comprehensive systematic review & SC: S & Grasping & L \\
		{\cite{52}} & 2024 & Parallel-continuum robots: a survey & PC: S-M & General & M \\
		{\cite{76}} & 2024 & Design, modeling, and evaluation of parallel continuum robots: a survey & PC: S-M-O & General & M \\
		{\cite{130}} & 2024 & Biomimetic soft-legged robotic locomotion, interactions and transitions in terrestrial, aquatic and multiple environments & SC: S-M-C & Locomotion & H \\
		{\cite{165}} & 2024 & Actuated Palms for Soft Robotic Hands: Review and Perspectives & SC: S & General & L \\
		{\cite{166}} & 2024 & Variable stiffness soft robotic gripper: design, development, and prospects & SC: S & General & L \\
		{\cite{167}} & 2025 & Dexterous hand towards intelligent manufacturing: A review of technologies, trends, and potential applications & SC: S & General & L \\
        {\cite{206}} & 2026 & Cooperative integrated surgical robots for high-performance targeted therapies & SC: O & Surgery & L \\
		\bottomrule
	\end{tabularx}
		\begin{tablenotes}
			\footnotesize 
			\item SC: Separated Collaboration; AC: Assistance Collaboration; PC: Parallel Collaboration
			\item S: Structure Design; M: Modeling; P: Motion Planning; C: Control Strategy; O: Other Contents
			\item H: High; M: Medium; L: Low (determined by the number of related publications involving CCRs)
		\end{tablenotes}
	\end{threeparttable}
\end{table*}

A structured literature search was conducted in the Web of Science and Scopus for publications over the past two decades, using
``continuum robot'' in combination with the descriptors
``collaborative,''  ``cooperative,'' ``multi-arm,'' ``coupled,'' and ``parallel.'' A total of 872 records were retrieved. Following duplicate removal, title and abstract screening, and full-text review, 175 studies were included in this survey. The main contributions of this survey are summarized as follows:

1) A closed-loop task-execution architecture for CCRs is constructed from an automation-system perspective, providing a system-level framework for organizing and analyzing CCR-related research.
2) The three collaboration modes of CCRs are systematically classified, and related research progress is systematically surveyed in terms of key technologies, including multi-CR task allocation, structural design, modeling, motion planning methods, control strategies, and sensing mechanisms.
3) Research opportunities in key technologies and system integration challenges for real-world deployment are discussed to support the further development of CCRs.

The rest of this survey is organized as follows. Section \ref{section_xz} presents a closed-loop task-execution architecture for CCRs. Section \ref{section2} defines the three collaborative modes of CCRs and introduces their structural designs. Section \ref{section3} summarizes two modeling strategies of CCRs and further discusses the formulation of coupling constraints. Section \ref{section4} reviews motion planning methods and control strategies for CCRs. Section \ref{section5} discusses current challenges and future opportunities. Finally, Section \ref{section6} concludes the survey.

\section{CLOSED-LOOP TASK-EXECUTION ARCHITECTURE OF CCRs}
\label{section_xz}
CCRs are task-oriented collaborative systems of inherently compliant CRs,
rather than mere collections of CRs. Inspired by layered autonomous-robot architectures \cite{198}, a closed-loop task-execution architecture for CCRs is presented from an automation-system perspective to characterize the system-level organization from task decision-making and coordinated execution to feedback-based adjustment (Fig. \ref{Fig_auto}).

\subsection{Decision level}
\begin{figure} [t] 
	\centering
	\includegraphics[scale=0.46]{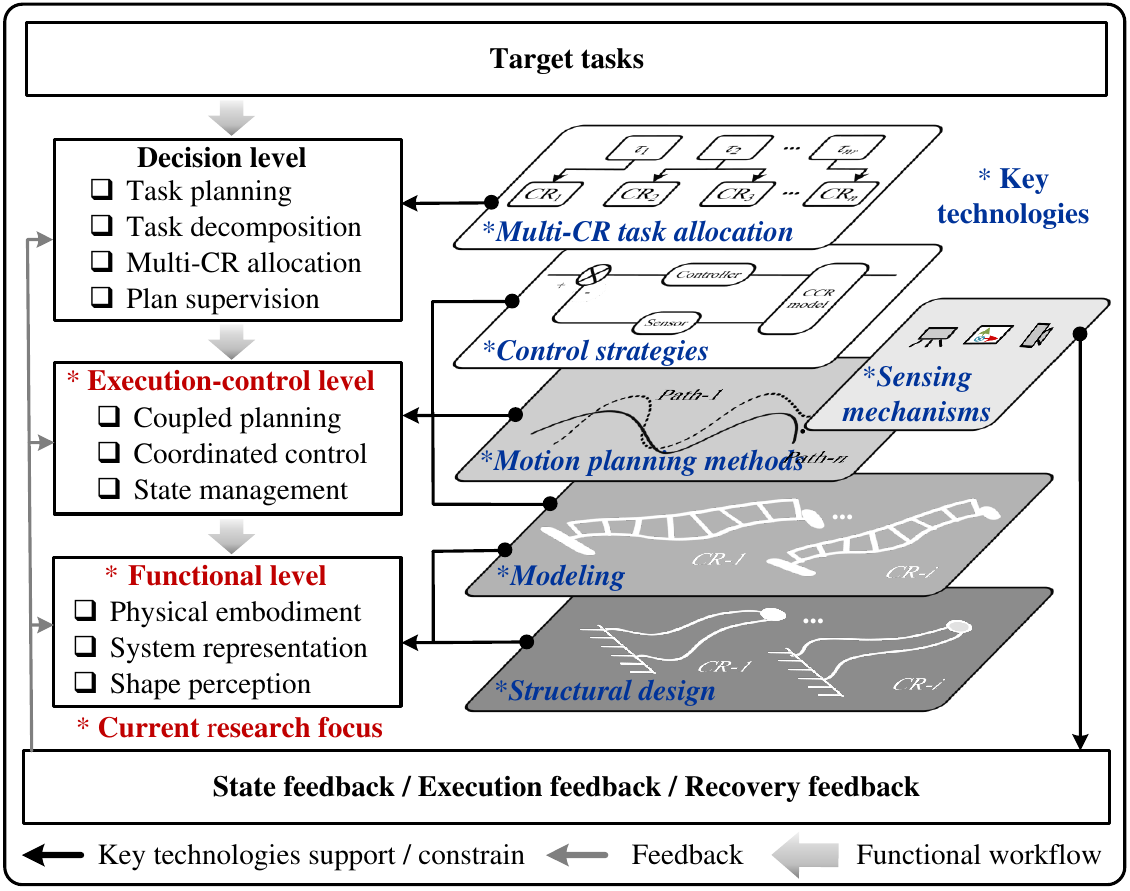} 
	\centering 
	\caption{Closed-loop task-execution architecture of CCRs from an automation-system perspective.}
	\label{Fig_auto} 
\end{figure}

The decision level converts the target task into executable multi-CR collaborative tasks through task planning, task decomposition, multi-CR allocation, and plan supervision (Fig. \ref{Fig_auto}). Task planning defines the execution workflow, task decomposition generates executable subtasks, multi-CR allocation assigns these subtasks to appropriate CRs or CR subsets, and plan supervision updates the task plan based on sensing mechanisms, triggering task reallocation when necessary.

Among these functions, task decomposition and allocation provide the key interface between target-task specification and the execution-control level \cite{199}. Let
$\mathcal{T}=\{\tau_1,\tau_2,\ldots,\tau_{n_r}\}$ denote the set of subtasks decomposed from the target task, and let $
\mathcal{R}=\{\mathrm{CR}_1,\mathrm{CR}_2,\ldots,\mathrm{CR}_n\}$
denote the set of CRs in the CCR system. Considering the multi-CR coupling characteristics of CCRs, the task-allocation mapping can be extended from the one-to-one mapping \(\mathcal{A}:\mathcal{T}\rightarrow\mathcal{R}\), commonly used in conventional multi-robot task allocation \cite{200,201}, to a subset-based cooperative allocation mapping
\begin{equation}
\mathcal{A}:\mathcal{T}\rightarrow 2^{\mathcal{R}},
\label{eq:task_allocation_mapping}
\end{equation}
where \(2^{\mathcal{R}}\) denotes the power set of \(\mathcal{R}\), namely the set of all possible CR subsets. For any subtask \(\tau_j\), the allocation result can be expressed as \(\mathcal{A}(\tau_j)=\mathcal{R}_j\), where \(\mathcal{R}_j\subseteq\mathcal{R}\) and \(\mathcal{R}_j\neq\emptyset\). Here, \(\mathcal{R}_j\) denotes the subset of CRs assigned to execute subtask \(\tau_j\). From an optimization perspective, task allocation can be abstracted as finding the optimal allocation under a set of constraints \cite{201}, so as to optimize the overall system performance \(\mathcal{U}_{\mathrm{sys}}(\mathcal{A})\) 
\begin{equation}
\begin{aligned}
\mathcal{A}^{*}
=&
\arg\max_{\mathcal{A}} \quad 
 \mathcal{U}_{\mathrm{sys}}(\mathcal{A}) \\
\mathrm{s.t.}\quad
& g_{\mathrm{str}}(\mathcal{A})\leq 0, \quad g_{\mathrm{mot}}(\mathcal{A})\leq 0, \quad g_{\mathrm{coup}}(\mathcal{A})\leq 0.
\end{aligned}
\label{eq:decision_allocation_optimization}
\end{equation}
where \(\mathcal{A}^{*}\) denotes the optimal task-allocation result, and \(\mathcal{U}_{\mathrm{sys}}(\mathcal{A})\) denotes the overall system performance associated with the allocation \(\mathcal{A}\).  
$g_{\mathrm{str}}(\mathcal{A})$, $g_{\mathrm{mot}}(\mathcal{A})$, and
$g_{\mathrm{coup}}(\mathcal{A})$ denote the structural feasibility,
motion feasibility, and coupling compatibility constraints, respectively.
Their feasibility boundaries are defined by configuration-dependent
structural limits, reachability and minimum-clearance requirements, and
prescribed tolerances on coupling residuals, respectively. Given the infinite DOFs and nonlinear deformation behavior of CRs
\cite{2,90}, a finite-dimensional parameterization of their configurations
can be used to render these constraints computationally tractable.
In particular, coupling compatibility can be assessed using residuals
associated with pose, velocity, and force/moment compatibility conditions,
as detailed in Section~\ref{section3}\textcolor{darkred}{-B}. When necessary, configuration-dependent nonlinear constraint functions, particularly coupling residuals, can be locally linearized around the current configuration using their Jacobians to facilitate feasibility screening of candidate task allocations. Notably, in complex task scenarios, task allocation often needs to be dynamically updated \cite{202}. For CCRs, this requirement becomes more pronounced because deformable-body behavior and inter-CR coupling may compromise the feasibility of predefined decisions. However, in existing CCR studies, such decision-making processes are still largely specified manually or guided by experience \cite{2,72,98}.

\subsection{Execution-control level}
The execution-control level transforms the task-allocation results generated at the decision level into executable multi-CR coordinated motions. This level mainly involves coupled planning, coordinated control, and state management (Fig. \ref{Fig_auto}). Coupled planning, supported by motion planning methods \cite{6,33,21}, transforms allocated subtasks into feasible paths/trajectories based on task objectives, environmental information, and system models. Coordinated control, implemented through control strategies \cite{9,10,91}, converts the planned motions into actuation commands while maintaining multi-CR synchronization, constraint satisfaction, and error compensation. State management, supported by sensing mechanisms \cite{1,44,61}, monitors task-execution and triggers local adjustment or subsequent replanning when necessary.
\subsection{Functional level}
The functional level supports the execution-control level by providing physical embodiment, system representation, and perception/state estimation  (Fig. \ref{Fig_auto}). The physical embodiment, realized through structural design \cite{20,16,66}, provides the structural basis for CCR task execution. System representation, established through modeling \cite{13,58,102}, describes CR deformation states and multi-CR coupling relationships, thereby providing computable system information for the execution-control level. Sensing mechanisms support the perception of CR shapes and the estimation of inter-CR relative states \cite{1,44,61}, providing feedback for closed-loop adjustment, execution monitoring, and task supervision.

\subsection{Cross-layer integration of key technologies}
Multi-CR task allocation, structural design, modeling, motion planning, sensing, and control are not independent, and their cross-layer integration jointly affects closed-loop task execution in CCRs. Structural design determines the coupling configuration and boundary conditions for modeling, while model complexity and modeling assumptions further affect the computational cost of motion planning and control performance. Meanwhile, sensing and processing latency may cause state feedback to lag behind actual execution, reducing planning responsiveness and limiting feasible control rates. Task allocation, in turn, accounts for structural feasibility, coupling compatibility, and motion feasibility, and is updated based on execution feedback.  Accordingly, closed-loop task execution in CCRs requires coordinated development across these key technologies. Given the limited research on multi-CR task allocation, the following sections focus on advances in the other key technologies.

\section{STRUCTURAL DESIGN FOR CCRs}
\label{section2}
From an automation-system perspective, structural design provides the physical basis for CCR performance and determines whether reliable closed-loop task execution can be achieved. Unlike single CRs, CCRs require the additional design of multiple CRs’ configuration layouts and interconnections to meet specific task requirements \cite{66}. In this section, three representative collaboration modes are defined, and their structural forms are presented and analyzed.

\subsection {Fundamentals of structural design}
As a fundamental premise, the design of CCRs follows the established design principles of single CRs \cite{18}. Accordingly, depending on task requirements, CCRs typically adopt actuation schemes such as concentric tube, tendon-driven, rod-driven, pneumatic and hydraulic, origami-inspired, magnetic, and hybrid actuation (Table~\ref{Tab_CCRModes}). Building upon these foundations, additional coupling-related design issues \cite{25} should be considered.

\subsection {Separated collaboration}
In the separated collaboration mode, $n \,(n \geq 2)$ CRs without direct mechanical linkage operate independently but in coordination to accomplish a shared task, thereby improving task adaptability (Fig. \ref{Fig_2}\textcolor{darkred}{(a)}) \cite{18,19,209}. By equipping individual CRs with different end-effectors, this mode enables cooperative tasks beyond the capability of a single CR, such as cutting \cite{13,30,38,53}, suturing \cite{23}, and repair \cite{2}. However, this tool diversity also places higher performance demands on the individual CRs.

To address the performance demands placed on individual CRs in this mode, various approaches have been proposed. One approach is parameter optimization, which can exploit performance potential while preserving the existing structural framework, for example by optimizing the CCR body using particle swarm optimization \cite{9}, the Nelder–Mead simplex algorithm \cite{19}, and the adaptive parameter grey wolf–cuckoo search algorithm \cite{53}. Another approach is to introduce additional structural designs to improve structural performance more directly. For example, load-bearing capacity can be enhanced through rolling-joint-based redundant-DOF constraint mechanisms \cite{31,38}, external locking cables \cite{24}, and variable-stiffness structures based on granular and layer jamming techniques \cite{46}; meanwhile, hybrid rigid--flexible structures \cite{3,22} and mixed actuation schemes \cite{14} can help mitigate inter-segment coupling effects. However, the performance gains achieved by parameter-optimization-based approaches are often limited, whereas additional structural configurations usually come at the cost of increased structural complexity, thereby raising the risk of collision. To better exploit their respective advantages, these two strategies can be combined at different stages of the design process to further enhance overall performance. Such improvements are often associated with greater system integration complexity.

As the structural complexity of individual CRs increases, interference avoidance becomes an equally important consideration. On the one hand, collisions among CRs can be mitigated by adjusting their angular arrangement \cite{35} and base pose \cite{90}. On the other hand, interference among internal components can be reduced by adopting a pyramid-type actuation structure (Fig. \ref{Fig_2}\textcolor{darkred}{(b)}) \cite{27,28}, with its design parameters further optimized through multi-objective optimization \cite{25,26}. However, in complex task scenarios, how to systematically determine the spatial layout and structural configuration according to task requirements so as to balance collision avoidance, accessibility, and task performance still deserves further investigation.

\begin{table*}[t]
\centering
\begin{threeparttable}
\caption{\textbf{Comparison of Collaborative Modes for CCRs: Actuation and Applications}}
\label{Tab_CCRModes}
\setlength{\tabcolsep}{3.5pt}
\renewcommand{\arraystretch}{1.12}
\begin{tabular}{
>{\centering\arraybackslash}m{0.070\textwidth}
>{\centering\arraybackslash}m{0.500\textwidth}
>{\centering\arraybackslash}m{0.130\textwidth}
>{\centering\arraybackslash}m{0.160\textwidth}
}
\toprule
\textbf{Mode} &
\textbf{Ref.} &
\textbf{Actuation} &
\textbf{Application} \\
\midrule

\multirow{7}{*}{SC}
& \makecell[c]{\cite{142}, \cite{18}, \cite{20}, \cite{11}, \cite{19}, \cite{30},
 \cite{23}, \cite{45}, \cite{28}, \cite{9}, \cite{29}}
& \makecell[c]{Concentric tube}
& SU \\
& \makecell[c]{\cite{2}, \cite{90}, \cite{81}, \cite{96}, \cite{13}, \cite{38},
\cite{53}, \cite{31}, \cite{24}, \cite{22}, \cite{35}, \cite{110}}
& Tendon
& \makecell[c]{SU-MA-GR-LO} \\
& \makecell[c]{\cite{135}, \cite{3}}
& Rod
& SU \\
& \makecell[c]{\cite{108}, \cite{111}, \cite{80}, \cite{69}, \cite{109},
\cite{102}, \cite{112}, \cite{115}}
& Pneumatic
& GR-LO \\
& \makecell[c]{\cite{107}}
& Hydraulic
& GR \\
& \makecell[c]{\cite{93}, \cite{113}}
& Origami
& GR \\
& \makecell[c]{\cite{89}, \cite{103}, \cite{46}, \cite{14}, \cite{104}, \cite{84}}
& Hybrid
& \makecell[c]{SU-GR-LO} \\
\midrule

\multirow{3}{*}{AC}
& \makecell[c]{\cite{60}, \cite{57}, \cite{66}, \cite{125}, \cite{16}, \cite{67}}
& Tendon
& \makecell[c]{SU-MA-LO} \\
& \makecell[c]{\cite{88}, \cite{161}}
& Magnetic
& SU \\
& \makecell[c]{\cite{70}}
& Hybrid
& MA \\[3pt]
\midrule

PC
& \makecell[c]{\cite{119}, \cite{73}, \cite{78}, \cite{121}, \cite{58}, \cite{77}, \cite{62}}
& Tendon
& SU-MA \\
\bottomrule
\end{tabular}

\begin{tablenotes}
\footnotesize
\item SC: Separated Collaboration; AC: Assistance Collaboration; PC: Parallel Collaboration
\item SU: Surgery; MA: Maintenance; GR: Grasping; LO: Locomotion
\end{tablenotes}
\end{threeparttable}
\end{table*}

In addition, to improve the spatial deployment capability of CCRs in complex environments, existing studies have mainly focused on guiding-structure design and integration with mobile bases. On the one hand, rigid guiding structures with axial channels (Fig. \ref{Fig_2}\textcolor{darkred}{(c)}) can direct multiple CRs to distal regions, thereby expanding the workspace in confined environments \cite{20,45}. Building on this, the guiding structure can be further designed as a modular form with replaceable CRs (Fig. \ref{Fig_2}\textcolor{darkred}{(d)}) \cite{28,90} to improve system versatility and maintainability; when the target path involves frequent bends, it can also be designed as a flexible CR (Fig. \ref{Fig_2}\textcolor{darkred}{(e)--(f)}) to enhance the path adaptability and dexterity of the CCR \cite{11,53,89,135}. On the other hand, integrating CCRs with mobile bases such as unmanned aerial vehicles \cite{112,162}, automated guided vehicles \cite{68}, industrial robots \cite{129,186,187}, or CCR-based legged robots (Fig. \ref{Fig_2}\textcolor{darkred}{(j)}) \cite{4,96,97,113,115,84,123,124,142} can further expand their task adaptability at the system level. These approaches, however, usually entail greater system-integration difficulty and overall complexity.

Building on the above studies of general structural design, this mode can be further developed into task-oriented structural implementations, among which object grasping is one of the most representative directions \cite{108,155}. In such cases, individual CRs are anchored to the same or different mobile/fixed bases and grasp the target object through coordinated deformation \cite{6,95}. In recent years, diverse structural forms have been developed, including variable-stiffness tendon-driven structures based on layer jamming \cite{104}, actuators composed of magnetically programmable soft materials \cite{105,106}, and pneumatic artificial muscle–driven structures \cite{69,109,108}. Integrated assemblies of these structures can form CCRs \cite{99, 142, 188, 189}. However, limitations still remain in grasping complex objects.
By mimicking the coiling and entangling functions of octopus tentacles, elephant trunks, and plants \cite{111,114}, bioinspired designs provide an effective solution to this problem. For instance, the design of multiple fluid-driven CRs enables stochastic-topology grasping (Fig. \ref{Fig_2}\textcolor{darkred}{(h)}), thereby eliminating the need for sophisticated perception and feedback strategies during the grasping process \cite{111}. Further, incorporating hard bulges on the CR surface allows for diverse grasping strategies \cite{112}, while tendon-driven logarithmic spiral–shaped simplified structures (Fig. \ref{Fig_2}\textcolor{darkred}{(i)}) improve controllability in grasping tasks \cite{81}. In addition, dexterous hands composed of CR-based fingers (Fig. \ref{Fig_2}\textcolor{darkred}{(j)}), which commonly adopt tendon-driven \cite{110}, pneumatic \cite{80,102}, hydraulic \cite{107}, dielectric-elastomer-based \cite{99}, origami-inspired \cite{93}, or hybrid actuation schemes \cite{103}, can also be regarded as a form of separated collaboration. However, in most dexterous hands, each finger mimics human anatomy through a finite number of discrete joints, placing them outside the scope of this survey. Comprehensive reviews are available in \cite{165,166,167}.

Separated collaboration is evolving from enhancing the performance of individual CRs to enabling multi-CR coordination, thereby improving task adaptability. However, its load-bearing capability remains relatively limited, and the risk of interference increases as task scenarios become more complex (Table~\ref{tab:collaboration_modes_comparison}). Mitigating these issues involves a trade-off between collaborative performance and structural complexity.

\begin{table*}[t]
\centering
\caption{\textbf{Structural and performance comparison of collaboration modes for CCRs}}
\label{tab:collaboration_modes_comparison}
\renewcommand{\arraystretch}{1}
\setlength{\tabcolsep}{3pt}
\setlength{\extrarowheight}{0pt}
\renewcommand{\tabularxcolumn}[1]{m{#1}}
\begin{threeparttable}
\begin{tabularx}{\textwidth}{
    >{\centering\arraybackslash}m{0.05\textwidth}
    >{\centering\arraybackslash}m{0.05\textwidth}
    >{\centering\arraybackslash}X
    >{\centering\arraybackslash}m{0.08\textwidth}
    >{\centering\arraybackslash}m{0.08\textwidth}
    >{\centering\arraybackslash}m{0.11\textwidth}
    >{\centering\arraybackslash}m{0.08\textwidth}
    >{\centering\arraybackslash}m{0.08\textwidth}
    >{\centering\arraybackslash}m{0.08\textwidth}
    >{\centering\arraybackslash}m{0.1\textwidth}
    >{\centering\arraybackslash}m{0.08\textwidth}
}
\toprule
\textbf{Mode} &
\shortstack[c]{\textbf{No.}\\\textbf{of CRs}} &
\shortstack[c]{\textbf{Physical}\\\textbf{coupling}} &
\shortstack[c]{\textbf{Connection}\\\textbf{type}} &
\shortstack[c]{\textbf{Role}\\\textbf{relationship}} &
\shortstack[c]{\textbf{System}\\\textbf{end-effector}} &
\shortstack[c]{\textbf{Integration}\\\textbf{complexity}} &
\shortstack[c]{\textbf{Load-}\\\textbf{bearing}} &
\shortstack[c]{\textbf{End-effector}\\\textbf{flexibility}} &
\shortstack[c]{\textbf{Interference}\\\textbf{risk}}  \\
\midrule
SC & $\geq 2$ & None & None & Peer/non-peer & Individual CR tips & L & L & M & H  \\
AC & $\geq 2$ & Side connection to the main arm & \shortstack[c]{Fixed/\\detachable} & Non-peer & Main arm tip & M & M & H & M \\
PC & $\geq 2$ & Distal connection to a moving platform & Fixed & Peer & Moving platform & H & H & L & L  \\
\bottomrule
\end{tabularx}

\begin{tablenotes}
\footnotesize
\item SC: Separated Collaboration; AC: Assistance Collaboration; PC: Parallel Collaboration
\item H: High; M: Medium; L: Low (The ratings provide a relative qualitative comparison among the three modes, assuming identical constituent-CR designs and intrinsic performance. Integration complexity is assessed based on the difficulty of implementing
the required inter-CR physical coupling; load-bearing capability, on the
degree of load sharing enabled by the coupling; end-effector
flexibility, on the reachable motion range of the system end-effector under
the corresponding coupling constraints; and interference risk, on the potential for inter-CR interference arising from relative motion.)
\end{tablenotes}
\end{threeparttable}
\end{table*}

\subsection {Assistance collaboration}
The assistance collaboration mode of CCRs refers to a configuration in which $n \,(n \geq 2)$ CRs cooperate with unequal roles to accomplish a task. In this mode, one or more CRs act as support arms, with their ends connected to the side of the main CR \cite{16}, thereby assisting the main CR in accomplishing the target task.

A representative configuration is to enhance the stiffness of the master arm through local auxiliary support. Inspired by the growth pattern of the Angel Oak Tree, a tendon-driven concentric-tube structure with lateral branches has been proposed, in which support arms extend laterally to support the motion of the continuously extending master arm (Fig. \ref{Fig_2}\textcolor{darkred}{(k)}) \cite{57}. However, this configuration remains limited in its adaptability to confined spaces.

To improve adaptability, this mode has evolved toward configurations with detachable connections. These configurations allow small-diameter master and support arms to enter through different access points and establish connections in the target region (Fig. \ref{Fig_2}\textcolor{darkred}{(l)}) \cite{16}. On this basis, the stiffness of the master arm can be tuned by varying the position and angle of the connection points \cite{66,126}. A key challenge in realizing such configurations lies in the design of detachable connections, which have been implemented using a shape-memory-alloy-based thermally triggered attachment/detachment mechanism (Fig. \ref{Fig_2}\textcolor{darkred}{(m)}) \cite{70} and an open-close multi-finger spherical gripper (Fig. \ref{Fig_2}\textcolor{darkred}{(n)}) \cite{125}. Although these designs provide more flexible support for the master arm, the trade-offs among connection reliability, response speed, and structural complexity still require systematic evaluation.

Building on detachable  connection, support in this mode has further evolved into multi-point or switchable architectures along the motion path. One approach adopts a slender master arm with support arms deployed at multiple locations along its path and laterally connected to it, thereby stabilizing the proximal region of the master arm and improving its compliance without increasing the overall size of the CR (Fig. \ref{Fig_2}\textcolor{darkred}{(o)}) \cite{60}. Another approach employs a magnetically actuated parallel casing structure, in which individual casings alternately harden and soften to achieve alternating follower motion (Fig. \ref{Fig_2}\textcolor{darkred}{(p)}) \cite{88,161}. However, the further application of such configurations remains constrained by system integration complexity and dynamic response capability.

Assistance collaboration is evolving from fixed local support toward reconfigurable support architectures, thereby improving the flexibility of the master arm and, to some extent, enhancing its local load-bearing capability (Table~\ref{tab:collaboration_modes_comparison}). However, the design of reliable connection and support structures still involves trade-offs, which remain obstacles to further development.

\subsection {Parallel collaboration}

The parallel collaboration mode of CCRs refers to a configuration in which $n \,(n \geq 2)$ peer CRs are symmetrically arranged in a parallel architecture, with their ends connected to a moving platform that serves as the CCR end-effector.

PCRs constructed using rods arranged in parallel \cite{71,74}, compliant joints \cite{75}, or pneumatic chambers \cite{120} as actuation elements can be considered the earliest form of the parallel collaborative configuration, with comprehensive summaries available in \cite{76} and \cite{52}. However, these actuation elements rely solely on passive deformation through axial extension or contraction under material constraints, essentially remaining within the scope of single CRs. To further enhance motion capabilities and stiffness characteristics, parallel collaborative configurations composed of CRs with active deformation capability \cite{58} have been proposed. 

Currently, representative studies focus on planar configurations either consisting of three tendon-driven CRs (Fig. \ref{Fig_2}\textcolor{darkred}{(q)}) or comprising three CRs combined with passive rigid links (Fig. \ref{Fig_2}\textcolor{darkred}{(r)}). The individual components are interconnected via spherical joints, enabling the moving platform to achieve 4 DOFs ($\vec{x}$, $\vec{y}$, $\vec{z}$, and $\widetilde{z}$), with high repeatability \cite{73,77,78}. By optimizing the parameters of the CRs, passive rigid links, and actuation layouts, the size and shape of the workspace can be adjusted to obtain configurations suited to specific applications \cite{62}. Although planar configurations can offer relatively simpler structural implementation, their limited workspace restricts the execution of more complex tasks. 

To overcome this limitation, parallel collaboration has been further extended to spatial architectures. Transitioning from planar to spatial arrangements of CRs increases the moving platform's DOFs to 6 ($\vec{x}$, $\vec{y}$, $\vec{z}$, $\widetilde{x}$, $\widetilde{y}$, and $\widetilde{z}$), thereby enhancing its motion capabilities (Fig. \ref{Fig_2}\textcolor{darkred}{(s)--(t)}) \cite{121,58}. In addition, the load-bearing capacity of the moving platform can be further enhanced either by increasing the number of parallel CRs (Fig. \ref{Fig_2}\textcolor{darkred}{(u)}) \cite{119} or by employing multiple constraint disks to form a distributed antagonistic configuration (Fig. \ref{Fig_2}\textcolor{darkred}{(v)}) \cite{192}. However, these configurations also increase structural complexity and place stricter demands on compact integration, assembly quality, and structural consistency.

Parallel collaboration is evolving from planar configurations to spatial configurations. Compared with the other two collaboration modes, this mode is more suitable for achieving high-load output, while its support relationships are more fixed, which may limit task adaptability and flexibility (Table~\ref{tab:collaboration_modes_comparison}).

\subsection {Summary}
It is worth noting that some configurations may exhibit features of multiple collaboration modes. In such ambiguous cases, classification should be based primarily on the organization of the end-effector. For example, although proximally multi-disk-constrained concentric-tube structures involve configurational coupling \cite{11}, the portions extending beyond the end disk still operate independently without fixed inter-arm connections and thus belong to separated collaboration. By contrast, fully multi-disk-constrained configurations \cite{192}, whose continuum units are distally connected to a common end disk serving as the system end-effector, should be classified as parallel collaboration. Detachable-connection configurations, although able to switch between separated and assistance collaboration \cite{70,125}, should still be regarded as assistance collaboration because their connection mechanism mainly provides auxiliary support to the master arm as the end-effector. Table~\ref{tab:collaboration_modes_comparison} further compares the characteristics of the three collaboration modes to clarify the classification criteria.

The three collaboration modes of CCRs represent distinct structural organization strategies. Separated collaboration preserves the structural independence of individual CRs, thereby reducing integration complexity, although limited load-bearing capability and inter-arm interference remain concerns; assistance collaboration enhances the master arm through local connection and support structures, improving local structural performance while placing higher demands on connection reliability and stability; parallel collaboration increases the load-bearing capability of the moving platform through relatively fixed coupling relationships, whereas such constraints may reduce structural flexibility (Table~\ref{tab:collaboration_modes_comparison}). These differences indicate that CCR structural design is a application-specific configuration design problem involving mode-specific structural trade-offs. However, current design processes remain highly dependent on designer expertise, making it difficult to systematically balance integration complexity, load-bearing capability, end-effector flexibility, and interference risk.

\begin{figure*} [t] 
	\centering
	\includegraphics[scale=0.5]{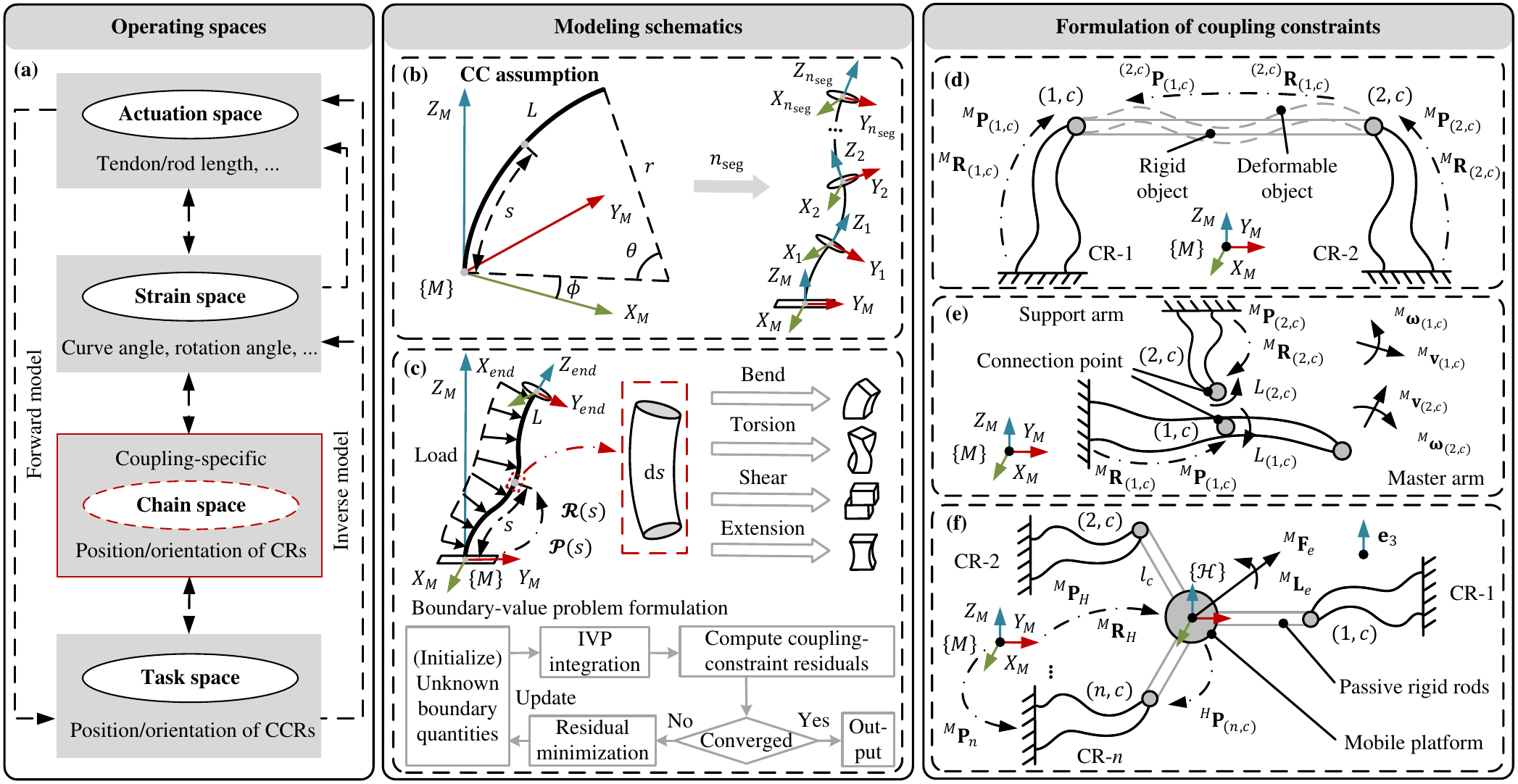} 
	\centering 
	\caption{Coupling modeling strategies for CCRs. (a) Operating spaces of coupling modeling strategies.
		(b) Schematic of the PCC model.
		(c) Schematic of the Cosserat rod theory.
		(d) Formulation of coupling constraints for separated collaboration.
		(e) Formulation of coupling constraints for assistance collaboration.
		(f) Formulation of coupling constraints for parallel collaboration.}
	\label{Fig_3} 
\end{figure*}

\section {MODELING FOR CCRs}
\label{section3}
From an automation-system perspective, modeling provides a mathematical bridge between the physical CCR system and the upper system layers. Building on the coupled structural design of CCRs, a mathematical model needs to be developed through parameterized structural features to incorporate kinematic states and mechanical coupling constraints among $n$ CRs, thereby ensuring coordinated motion of the CCR. In this section, two common modeling strategies for CCRs are introduced, followed by a description and summary of the methods for formulating coupling constraints.

\subsection {Modeling strategies}
Due to the structural complexity of CCRs, two modeling strategies are commonly employed, which include independent modeling and coupled modeling. 

In the independent modeling strategy, each CR within the CCRs is modeled separately, resulting in a mathematical model that is the “sum” of multiple independent CR models  \cite{13,35,53}. In contrast, the coupled modeling strategy builds upon independent modeling by incorporating geometric, velocity, and force constraints among the CRs to form a closed-loop kinematic chain, thereby constraining their relative motions. In this case, each CR’s workspace can be regarded as a chain space, while the overall workspace of the CCR constitutes the final task space, as shown in Fig. \ref{Fig_3}\textcolor{darkred}{(a)}. At the strategy level, independent modeling is generally more suitable when each CR operates independently, a condition more commonly encountered in separated collaboration \cite{32,37}. However, its robustness deteriorates once inter-CR coupling becomes significant, potentially leading to antagonism or collision. In contrast, coupled modeling explicitly incorporates coupling constraints and is therefore more suitable for strongly coupled scenarios, particularly in assistance \cite{17} and parallel collaboration \cite{59}, although its robustness remains contingent on the validity of those constraints and may degrade when constraint errors, parameter uncertainties, or unmodeled interactions become significant.

In both strategies, implementations are commonly based on the constant curvature (CC) assumption in geometric models and the Cosserat rod theory in physical models. The pseudo-rigid-body model (PRB) \cite{77}, another geometric modeling method, is less frequently employed.

\subsubsection*{1) CC assumption}

As an idealized deformation paradigm for CRs, the CC assumption considers each segment of a deformed CR to be representable by a finite set of arc parameters \cite{58,102}. Let \( L \) denote the total length of the CR. For any point \( s \in [0, L] \), the homogeneous transformation matrix \( \mathbf{T}_t(\kappa, \phi, s) \in \mathbb{R}^{4 \times 4} \) of a single segment is given by \cite{55,62}
\begin{equation}
\mathbf{T}_t(\kappa, \phi, s) =
\begin{bmatrix}
	c\phi \, c\theta & -s\theta & c\phi \, s\theta & r \, c\phi (1 - c\theta) \\
	s\phi \, c\theta & c\phi    & s\phi \, s\theta & r \, s\phi (1 - c\theta) \\
	-s\theta         & 0        & c\theta          & r \, s\theta \\
	0                & 0        & 0                & 1
\end{bmatrix},
\end{equation}
where \( \kappa \) is the curvature, \( \theta = \kappa s \) is the bending angle, \( r = 1/\kappa \) is the radius of curvature, and \( \phi \) is the rotation angle about the base \( Z \)-axis. For notational simplicity, define \( c\phi = \cos\phi \), \( s\phi = \sin\phi \), \( c\theta = \cos\theta \), and \( s\theta = \sin\theta \).

Based on the CC assumption, the model can be further extended to piecewise constant curvature (PCC) by stacking multiple segments, enabling the modeling of multi-segment CRs (Fig. \ref{Fig_3}\textcolor{darkred}{(b)}). In PCC, the end-effector pose transformation matrix $\mathbf{T}_{\text{end}}$ for a CR with $n_{\text{seg}}$ segments can be obtained through chain multiplication \cite{13,35,95}
\begin{equation}
\mathbf{T}_{\text{end}} = \prod_{i_{\text{seg}}=1}^{n_{\text{seg}}} \mathbf{T}_{i_{\text{seg}}} .
\end{equation}

\subsubsection*{2) Cosserat rod theory}

In the base coordinate frame, the position and orientation of an arbitrary point \( s \in [0, L] \) along the CR can be described by a position vector field \( \mathbf{P}(s) \in \mathbb{R}^3 \) and an orientation matrix field \( \mathbf{R}(s) \in \mathrm{SO}(3) \). Additionally, the robot is subjected to an internal force \( \mathbf{F}_{\text{in}}(s) \in \mathbb{R}^3 \), an internal moment \( \mathbf{L}_{\text{in}}(s) \in \mathbb{R}^3 \), a distributed external force \( \mathbf{F}(s) \in \mathbb{R}^3 \), and an external moment \( \mathbf{L}(s) \in \mathbb{R}^3 \) (Fig.~\ref{Fig_3}\textcolor{darkred}{(c)}).

Based on this parameterization, the balance equations at \( s \) are established under the Newton–Euler framework as follows \cite{29, 11, 17, 67}
\begin{equation}
	\begin{aligned}
		\frac{\partial \mathbf{P}(s)}{\partial s} &= \mathbf{K}_{\mathrm{lin}}(s), \,
		\frac{\partial \mathbf{R}(s)}{\partial s} = \hat{\mathbf{K}}_{\mathrm{cur}}(s)\, \mathbf{R}(s), \\
		\frac{\partial \mathbf{F}_{\text{in}}(s)}{\partial s} &= -\mathbf{F}(s), 
		\frac{\partial \mathbf{L}_{\text{in}}(s)}{\partial s} = -\mathbf{L}(s) - \mathbf{K}_{\mathrm{lin}}(s) \times \mathbf{F}_{\text{in}}(s),
	\end{aligned}
\end{equation}
where \( \mathbf{K}_{\mathrm{cur}} \in \mathbb{R}^3 \) and \( \mathbf{K}_{\mathrm{lin}}(s) \in \mathbb{R}^3 \) represent the curvature and linear strain of the CR, respectively, and \( \hat{\mathbf{K}}_{\mathrm{cur}}(s) \) denotes the skew-symmetric matrix of \( \mathbf{K}_{\mathrm{cur}}(s) \) \cite{66}. 

At the method level, CC-based models generally maintain better robustness when external loading is weak and deformation remains close to idealized curvature assumptions. By contrast, in assistance and parallel collaboration, where loading and coupling effects are often more pronounced, Cosserat rod models can accommodate richer deformation and loading effects under more complex conditions. Nevertheless, their robustness is also more dependent on the accurate specification of material parameters and external loads, making them more sensitive to modeling uncertainty.

\begin{table*}[t]
\centering
\setlength{\tabcolsep}{3pt}
\renewcommand{\arraystretch}{1}
\setlength{\extrarowheight}{0pt}
\renewcommand{\tabularxcolumn}[1]{m{#1}}
\caption{Comparison of modeling methods for CCRs}
\label{Tab_3}
\begin{tabularx}{\textwidth}{
>{\centering\arraybackslash}m{0.05\textwidth}
>{\centering\arraybackslash}m{0.05\textwidth}
>{\centering\arraybackslash}m{0.09\textwidth}
>{\centering\arraybackslash}m{0.08\textwidth}
>{\centering\arraybackslash}m{0.09\textwidth}
>{\centering\arraybackslash}m{0.06\textwidth}
>{\centering\arraybackslash}X
>{\centering\arraybackslash}m{0.10\textwidth}
>{\centering\arraybackslash}m{0.09\textwidth}
>{\centering\arraybackslash}m{0.06\textwidth}
>{\centering\arraybackslash}m{0.06\textwidth}
}
\toprule
\textbf{Mode} &
\textbf{Ref.} &
\shortstack[c]{\textbf{Modeling}\\\textbf{methods}} &
\shortstack[c]{\textbf{Constraint}\\\textbf{types}} &
\shortstack[c]{\textbf{Model}\\\textbf{type}} &
\shortstack[c]{\textbf{Constraint}\\\textbf{extensibility}} & \shortstack[c]{\textbf{I. K. solution}}
 &
\shortstack[c]{\textbf{Computational}\\\textbf{cost}} &
\shortstack[c]{\textbf{Numerical}\\\textbf{robustness}} &
$\boldsymbol{\varepsilon}_{\mathrm{pos}}$ &
$\boldsymbol{\varepsilon}_{\mathrm{ori}}$ \\
\midrule

\multirow[c]{8}{*}{SC}
& \cite{13}  & PCC          & None         & Kinematic    & NA & Numerical & NR        & $\mathrm{H}$                          & 0.463\% & NR     \\
& \cite{35}  & CC           & None         & Kinematic    & NA & Analytical & NR & $\mathrm{H}$                    & NR       & NR     \\
& \cite{53}  & PCC          & None         & Kinematic    & NA & NR                             & NR       & NR                         & NR & NR      \\
& \cite{72}  & PCC          & None         & Kinematic    & NA & Numerical & NR & $\mathrm{H}$                & 0.425\% & NR      \\
& \cite{95}  & PCC          & None        & Kinematic    & NA & Numerical                        & NR       & $\mathrm{H}$                          & 0.333\% & NR      \\
& \cite{33}  & CC           & GC         & Kinematic    & $\mathrm{M}$  & Semi-analytical & NR     & $\mathrm{H}$   & 1.200\% & NR \\
& \cite{67}  & Cosserat rod & GC-FC-VC   & Dynamic      & $\mathrm{H}$  & Numerical & NR & $\mathrm{L}$ & 3.950\% & NR      \\
& \cite{98}  & Cosserat rod & GC-FC      & Kinetostatic & $\mathrm{H}$  & \shortstack[c]{Numerical} & 0.100\,s & $\mathrm{M}$ & 1.040\% & NR      \\

\midrule

\multirow[c]{4}{*}{AC}
& \cite{11}  & Cosserat rod & GC-FC      & Kinetostatic & $\mathrm{H}$  & \shortstack[c]{Numerical} & 150.000\,s & $\mathrm{L}$ & 1.960\% & 2.133\% \\
& \cite{60}  & Cosserat rod & GC-FC      & Kinetostatic & $\mathrm{H}$  & \shortstack[c]{Numerical} & NR        & $\mathrm{L}$ & 2.168\% & NR      \\
& \cite{66}  & Cosserat rod & GC-FC-VC   & Dynamic      & $\mathrm{H}$  & Numerical & NR & $\mathrm{L}$ & 2.900\% & NR     \\
& \cite{17}  & Cosserat rod & GC-FC      & Kinetostatic & $\mathrm{H}$  & Numerical & 0.280\,ms & $\mathrm{M}$ & 0.180\% & 1.783\% \\

\midrule

\multirow[c]{4}{*}{PC}
& \cite{77}  & CC/PRB       & GC         & Kinematic    & $\mathrm{L}$  & Numerical & NR        & $\mathrm{H}$ & NR       & NR      \\
& \cite{62}  & CC           & GC         & Kinematic    & $\mathrm{L}$  & \shortstack[c]{Numerical} & NR & $\mathrm{H}$ & 1.400\% & 0.350\% \\
& \cite{73}  & CC           & GC         & Kinematic    & $\mathrm{L}$  & NR & NR & NR & 1.522\% & NR      \\
& \cite{59}  & Cosserat rod & GC-FC      & Kinetostatic & $\mathrm{H}$  & \shortstack[c]{Numerical} & 0.100\,s & $\mathrm{M}$ & 3.379\% & 1.973\% \\

\bottomrule
\end{tabularx}
\begin{tablenotes}
\footnotesize
\item H: High; M: Medium; L: Low (The ratings provide a relative qualitative comparison among the modeling methods. Constraint extensibility is qualitatively assessed based on the ability of the formulation to accommodate different physical coupling conditions: H denotes formulations that can further incorporate force/moment equilibrium and dynamic coupling; M denotes primarily geometric formulations applicable to broader scenarios when load- and interaction-induced effects need not be represented; and L denotes formulations largely limited to predefined geometric closure constraints. The numerical-robustness ratings are based on solution dimensionality, sensitivity to initial guesses, conditioning near singular configurations, the need for repeated integration and boundary-variable updates, and numerical stabilization strategies. H denotes low-dimensional formulations that generally avoid repeated integration and boundary updates and thus exhibit higher overall robustness; M denotes iterative formulations involving nonlinear iteration or repeated integration but enhanced by stabilization strategies such as nondimensionalization or null-space projection; and L denotes strongly coupled nonlinear formulations that are more sensitive to initial guesses and require repeated integration and boundary-variable updates.)
\item $\boldsymbol{\varepsilon}_{\mathrm{pos}}$: Relative mean position error (Defined as the actual mean position error divided by the length of the corresponding CR)
\item $\boldsymbol{\varepsilon}_{\mathrm{ori}}$: Relative mean orientation error (Defined as the actual mean angle error divided by $\pi$ rad)
\item 
NA: Not applicable; NR: Not reported
\item Note: Computational costs and errors depend on study-specific conditions and are therefore not used for direct quantitative comparisons between methods
\end{tablenotes}
\end{table*}

\subsection {Formulation of coupling constraints}

To achieve coordinated motion among CRs, coupled modeling is generally more appropriate. The modeling formulations for the three collaboration modes are presented below.

\subsubsection{Separated collaboration}

In separated collaboration, CRs operate without direct mechanical connections. When a task requires a fixed relative pose between CR ends, geometric constraints can be imposed at the corresponding end points \((1,c)\) and \((2,c)\) (Fig. \ref{Fig_3}\textcolor{darkred}{(d)}) \cite{6,68,33}
\begin{equation}
	\begin{aligned}
	{}^{M}\mathbf{P}_{(1,c)} - {}^{M}\mathbf{P}_{(2,c)} - {}^{M}\mathbf{R}_{(2,c)}\,{}^{(2,c)}\mathbf{P}_{(1,c)} &= \mathbf{0}, \\
		{}^{M}\mathbf{R}_{(1,c)} - {}^{M}\mathbf{R}_{(2,c)}\,{}^{(2,c)}\mathbf{R}_{(1,c)} &= \mathbf{0},
	\end{aligned}
\end{equation}
where \(\{M\}\) denotes the global frame, \(^{A}\mathbf{P}_B \in \mathbb{R}^3\) is the position vector of point \(B\) expressed in frame \(\{A\}\), and \(^{A}\mathbf{R}_B \in \mathrm{SO}(3)\) is the rotation matrix of the local frame at point \(B\) with respect to \(\{A\}\). This approach can concisely describe the relative poses of individual CRs. However, when multiple CRs grasp an object, pose constraints alone are insufficient, as contact forces/moments must also be considered \cite{6,67}
\begin{equation}
{}^{M}\mathbf{F}_{(i,c)} + {}^{M}\mathbf{F}_{(i,vc)} = \mathbf{0}, \quad
{}^{M}\mathbf{L}_{(i,c)} + {}^{M}\mathbf{L}_{(i,vc)} = \mathbf{0},
\end{equation}
where ${}^{M}\mathbf{F}_{(i,c)}$ and ${}^{M}\mathbf{L}_{(i,c)}$ denote the interaction force and moment acting on CR-$i$, respectively, and ${}^{M}\mathbf{F}_{(i,vc)}$ and ${}^{M}\mathbf{L}_{(i,vc)}$ denote the corresponding force and moment acting on the grasped object. When the grasped object is deformable, it can be modeled as a flexible rod (Fig. \ref{Fig_3}\textcolor{darkred}{(d)}) to account for its deformation and the resulting mechanical coupling among CRs \cite{67,98}. On this basis, velocity constraints can be imposed at the connection point to capture the dynamic response \cite{67}
\begin{equation}
{}^{M}\mathbf{v}_{(i,c)} = {}^{M}\mathbf{v}_{(i,vc)}, \quad {}^{M}\boldsymbol{\omega}_{(i,c)} = {}^{M}\boldsymbol{\omega}_{(i,vc)}.
	\label{eq:velocity_constraint}
\end{equation}
This may further extend the applicability of CCRs to high-speed motion scenarios. Nevertheless, the accuracy of such approaches still depends on the validity of the link abstraction and the accurate identification of object mechanical parameters, which may in turn limit their generalizability.

\subsubsection{Assistance collaboration}
In assistance collaboration, the distal end of the supporting arm CR-2 is physically connected to the lateral side of the main arm CR-1 (Fig. \ref{Fig_3}\textcolor{darkred}{(e)}). The local geometric constraints can still be formulated in a manner analogous to those in separated collaboration. The difference is that, owing to the direct connection between the two CRs, \( {}^{(2,c)}\mathbf{R}_{(1,c)} \) in (6) reduces to the identity matrix and \( {}^{(2,c)}\mathbf{P}_{(1,c)} \) reduces to the zero vector \cite{17}. Moreover, because of this direct connection, geometric constraints alone are usually insufficient, and force/moment constraints are also required for accurate characterization of the coupled deformation. These constraints take a form similar to that in (7), where \({}^{M}\textbf{F}_{(i,vc)}={}^{M}\textbf{F}_{(i+1,c)}\) and \({}^{M}\textbf{L}_{(i,vc)}={}^{M}\textbf{L}_{(i+1,c)}\) \cite{60,66}. Velocity constraints can be further introduced in a form analogous to (8), where \({}^{M}\textbf{v}_{(i,vc)}={}^{M}\textbf{v}_{(i+1,c)}\) and \({}^{M}\boldsymbol{\omega}_{(i,vc)}={}^{M}\boldsymbol{\omega}_{(i+1,c)}\) \cite{66}. 

Although these formulations effectively establish the relationships at the connection point, they place higher demands on the performance of the connecting mechanism, especially for detachable connection designs \cite{70}. Inadequate connecting performance may give rise to antagonistic interaction, whereas incorporating an appropriate tolerance into the constraint formulation may help mitigate this risk.

\subsubsection{Parallel collaboration}
In the parallel collaboration mode, the \(n\) CRs in the CCR are jointly connected to a mobile platform, and the backbone of each can be modeled as either a flexible rod or a combination of flexible and passive rigid rods \cite{62}. As the number of connection points increases, the system equilibrium shifts from the connection points to the center of the mobile platform (Fig.~\ref{Fig_3}\textcolor{darkred}{(f)}).

Based on the equilibrium point at the platform center, a closed-loop kinematic chain can be established for each CR, resulting in \(i\) position constraints (note that due to the use of spherical joint couplings, orientation constraints at the connection points are not required) \cite{73,77,78}
\begin{equation}
	\begin{aligned}
{}^{M}\mathbf{P}_H + {}^{M}\mathbf{R}_H \cdot {}^{H}\mathbf{P}_{(i,c)} &- {}^{M}\mathbf{P}_i - {}^{M}\mathbf{R}_H \cdot L_c \cdot \mathbf{e}_3 = \mathbf{0},\\
	& i = 1,2,\ldots,n \, ,
		\end{aligned}
\end{equation}
where \(\{H\}\) denotes the frame at the center of the mobile platform, \( {}^{H}\mathbf{P}_{(i,c)} \in \mathbb{R}^3 \) is the offset vector of the \(i\)-th connection point in frame \(\{H\}\), \( {}^{M}\mathbf{P}_i \in \mathbb{R}^3 \) is the end position of the \(i\)-th CR in the global frame \(\{M\}\), \( L_c \) is the length of the passive rigid rod, and \( \mathbf{e}_3 = [0,0,1]^\mathbf{T} \) represents the unit vector along the local z-axis of the CR. These geometric constraints are typically formulated based on geometric modeling \cite{73,77,78}. However, owing to multiple fixed connection points, interactions among CRs are generally more pronounced than in the other two collaboration modes. Therefore, a more suitable formulation may require physics-based modeling beyond purely geometric constraints, upon which global force/moment constraints can be further established at the center of the mobile platform \cite{59}.
\begin{equation}
	\begin{aligned}
{}^{M}\mathbf{F}_e - \sum_{i=1}^{n} \left( {}^{M}\mathbf{F}_{\text{in}, i} - \sum_k {}^{M}\mathbf{F}_{(k,i)} \right) &= \mathbf{0},\\
		{}^{M}\mathbf{L}_e - \sum_{i=1}^{n} \left( \mathbf{r}_i \times \left( {}^{M}\mathbf{F}_{\text{in}, i} - \sum_k {}^{M}\mathbf{F}_{(k,i)} \right) \right) &= \mathbf{0},
	\end{aligned}
\end{equation}
where \( {}^{M}\mathbf{F} \in \mathbb{R}^3 \) and \( {}^{M}\mathbf{L} \in \mathbb{R}^3 \) denote the external force and moment acting at the platform center. \( {}^{M}\mathbf{F}_{\text{in}, i} \) is the internal force at the end of the \(i\)-th CR, and \( {}^{M}\mathbf{F}_{(k,i)} \) represents the tensile force of the \(k\)-th tendon in the \(i\)-th CR, typically with \( k \in \{1,2,3,4\} \). The moment arm vector is given by \( \mathbf{r}_i = {}^{M}\mathbf{R}_H \cdot {}^{H}\mathbf{P}_{(i,c)} \in \mathbb{R}^3 \). This formulation improves scalability as the number of CRs increases, while inevitably increasing modeling complexity.

\subsubsection{Properties of coupling-constraint formulations}
After presenting the representative coupling-constraint formulations, their common characteristics can be analyzed. Among these characteristics, scalability is particularly important. Existing CC/PCC-based studies primarily use (6) and (9) to impose geometric closure constraints \cite{73,62}. Their scalability is limited because these constraints cannot readily be extended to more complex physical coupling. They remain applicable to a broader range of scenarios when load- and interaction-induced effects need not be represented \cite{33}. By contrast, Cosserat-rod-based formulations can further incorporate force/moment equilibrium in (7) and (10) and dynamic coupling in (8) \cite{17,59}, thereby supporting a broader range of physical interactions and offering greater scalability (Table~\ref{Tab_3}).

Another property is configuration singularity induced by coupling constraints. Related studies remain limited, possibly because current CCR research typically involves only a small number of CRs. In parallel collaboration, singularity analysis can often draw on methods developed for rigid parallel robots \cite{77,59}. However, in separated and assistance collaboration, coupling constraints established through virtual links \cite{98} or direct connection points \cite{30} between CRs may likewise introduce configuration singularities. Therefore, singularity analysis for CCRs, particularly the mechanisms of singularity formation and the corresponding criteria under different collaboration modes, still deserves further investigation.

\subsection {Numerical solution of the coupled model}
Coupled models involve additional coupling constraints, thereby making it more difficult to achieve high accuracy, computational efficiency, and numerical robustness in numerical solution, particularly for inverse problems.

\subsubsection {CC/PCC-based model}
CC/PCC-based models reduce CCR deformation to low-dimensional curvature variables. For numerical solution, one category casts (9) as a residual minimization problem \cite{62,73}, enabling multiple constraints to be handled within a unified framework, although the computational efficiency is often sensitive to the initial guess and solver settings. Another category differentiates (9) to derive a locally linear velocity relation and iteratively updates the configuration variables using the resulting Jacobian \cite{77}. However, near singular configurations, the inverse Jacobian or velocity transmission matrix may become ill-conditioned, reducing numerical robustness. Semi-analytical models can also be established by exploiting predefined relative geometric relationships among CRs described in (6) \cite{33}.

Although the computational cost of these methods has not yet been sufficiently quantified in the literature, CC/PCC-based models generally offer higher solution efficiency and better numerical stability than Cosserat-rod-based models, making them more suitable for real-time control. Under specific task conditions, they can achieve satisfactory accuracy (Table~\ref{Tab_3}). However, such accuracy relies on geometric assumptions, whose validity may deteriorate under strong coupling or external loading, thereby limiting model accuracy and applicability.

\subsubsection {Cosserat-rod-based model}
Cosserat rod theory typically describes a CR as a set of ordinary differential equations (ODEs) defined along arc length \cite{59,169}. For a single CR, the problem can usually be formulated as an initial-value problem and solved by numerical integration \cite{168}. In coupled modeling, the corresponding constraints residuals in (6)–(10) often convert the problem into a boundary-value problem.

For boundary-value problems, a common solution method is to perform multi-segment/multi-rod integration and update the unknown boundary quantities using a shooting method, while treating each subproblem as an initial-value integration at each iteration \cite{11,61,67}. On this basis, the corresponding constraints in (6)–(10) can be recast as residuals and minimized \cite{11,60} using numerical optimization methods such as sequential quadratic programming \cite{11}, the Levenberg--Marquardt algorithm \cite{66}, and the dogleg trust-region algorithm \cite{59} (Fig. \ref{Fig_3}\textcolor{darkred}{(c)}). When dynamic effects are further considered, this procedure is typically implemented within a time-discretization framework \cite{66,67}. Such methods provide a unified way to handle complex coupling constraints and continuum deformation. However, different initial guesses may lead to convergence to different configurations, indicating limited numerical robustness. Moreover, because the solution procedure repeatedly integrates the governing equations while updating boundary variables, the computational efficiency is low, and real-time control remains difficult to achieve directly. 

To alleviate these issues, existing studies have employed variable nondimensionalization \cite{17} and null-space projection \cite{98} to improve numerical robustness, reduced the number of unknowns through backward-forward integration \cite{17}, and used proper initialization together with recursive solution based on the previously computed solution \cite{59} to improve computational efficiency. Despite these efforts, the low computational efficiency and limited numerical robustness associated with repeated integration have not yet been fundamentally resolved. Although reports of computational cost remain limited, some representative methods have reduced the computational cost to the millisecond-level \cite{17,59,98} (Table~\ref{Tab_3}), suggesting that real-time control may be feasible in slow-motion scenarios such as surgical manipulation. However, further improvements in computational efficiency are still needed for applications requiring higher control frequencies. Moreover, when coupling is strong or external loads are non-negligible, Cosserat-rod-based models provide higher physical fidelity and therefore have greater potential to achieve higher accuracy than CC/PCC-based models. However, as the number and types of coupling constraints increase, the effects of constraint redundancy, inconsistency, and parameter errors may be further amplified, thereby limiting the attainable accuracy.

The validity of coupling constraints can be assessed in terms of numerical consistency and physical validity. Numerically, the residuals associated with the relevant constraints in (6)–(10) for a given collaboration mode can be examined to determine whether they converge below prescribed tolerances \cite{66}. Physically, the predicted inter-CR coupling states, such as relative poses and interaction forces/moments, can be compared with experimental measurements \cite{98}. Notably, small constraint residuals only indicate that the numerical solution satisfies the formulated equations, whereas the physical validity of the coupling constraints must be verified experimentally.

\subsection{Summary}
When the motion coordination \cite{33} and mechanical interactions \cite{59} among CRs need to be captured, coupled modeling is generally more appropriate. Across the three CCR collaboration modes, coupled models mainly differ in how constraints are organized. In separated collaboration, coupling is usually introduced through abstracted connection relationships \cite{98}, offering high flexibility, with modeling accuracy depending on the validity of the abstraction and the associated mechanical parameters. In assistance collaboration, constraints are mainly imposed at the connection points \cite{17}, enabling a more faithful description of local support and load transfer, with greater sensitivity to connection conditions and connection reliability, which may affect numerical robustness. In parallel collaboration, coupling is typically formulated as global constraints with respect to the moving platform center \cite{59}, improving closed-loop consistency and constraint scalability at the cost of reduced computational efficiency and a higher risk of constraint-induced singularities. Therefore, the trade-off between the physical fidelity of inter-CR interaction modeling and computational efficiency remains an important challenge.

\section{MOTION PLANNING, CONTROL, AND SENSING MECHANISMS FOR CCRs}
\label{section4}
From an automation-system perspective, motion planning and control should be coordinated at the execution-control level and supported by sensing mechanisms and modeling to achieve precise automated task execution while reducing collision and failure risks.

\begin{figure*} [t] 
	\centering
	\includegraphics[scale=0.48]{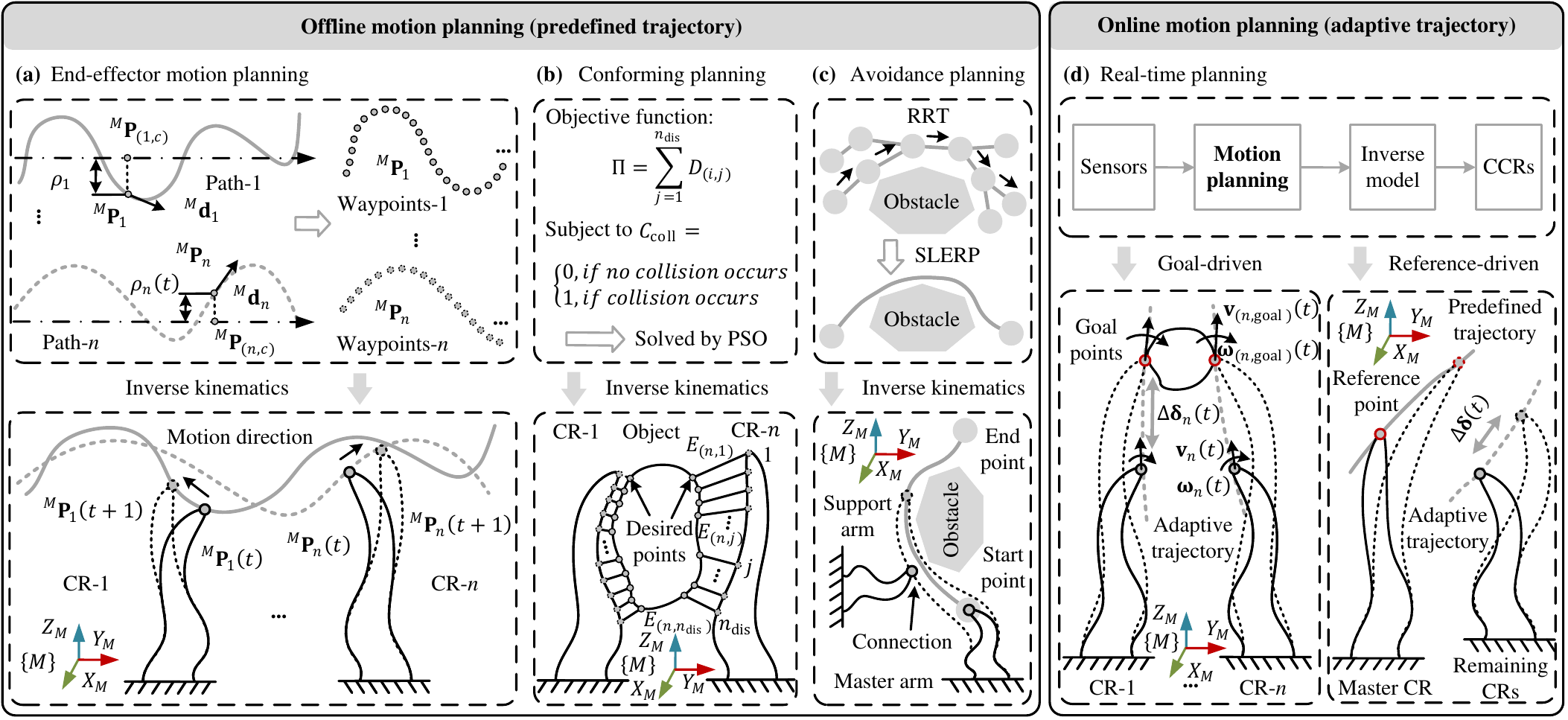} 
	\centering 
	\caption{Motion planning methods for CCRs. (a) End-effector planning. (b) Conforming planning. (c) Avoidance planning. (d) Real-time planning. }
	\label{Fig_4} 
\end{figure*}

\subsection{Motion planning for CCRs}
Defining an appropriate motion path serves as a guideline for accomplishing the target task. In current research, motion planning methods can be broadly classified into online and offline approaches.

\subsubsection*{1) Offline motion planning}
This approach refers to pre-planning the motion paths of each CR within the CCR workspace according to the task requirements before the target task begins, without further updates during execution. In this approach, the motion path ${^{M}\mathbf{P}}_{i} = [x_{i}, y_{i}, z_{i}]^\mathbf{T} $ can be defined by specifying the functional forms of a baseline and a pose offset (Fig.~\ref{Fig_4}\textcolor{darkred}{(a)}) \cite{72}
\begin{equation}
	{^{M}\mathbf{P}}_{i} = {^{M}\mathbf{P}}_{(i,c)} + \rho_{i} {^{M}\mathbf{d}}_{i},
\end{equation}
where 
\({^{M}\mathbf{P}}_{(i,c)} = [x_{(i,c)}, y_{(i,c)}, z_{(i,c)}]^\mathbf{T}\)  
denotes the position of the reference center point of the \(i\)-th path corresponding to the CR-\(i\),  
\(\rho_{i}\) is the offset radius of the path,  
and \({^{M}\mathbf{d}}_{i}\in \mathbb{R}^3\) is the direction vector of the path. As a simplified approach, it is more common in the three collaboration modes of CCRs to directly specify discrete waypoints ${^{M}\mathbf{P}}_{i}$ of the  CR-\(i\) end-effectors contained in the CCR \cite{96}, \cite{9}, \cite{8},   \cite{132}. Based on this, adding time constraints $t$ enables the realization of offline trajectory planning \cite{106}.
Although this approach is intuitive and easy to implement, it mainly specifies end-effector motion through a finite set of key poses, while giving limited consideration to task-specific interaction requirements.

In separated collaboration, interaction between the CRs and the environment is typically considered. Accordingly, grasp-oriented planning approaches for CCRs have been investigated, with \cite{82} proposing taxonomies and quantitative metrics for grasp quality, and \cite{83} further identifying optimal grasp configurations and contact points, thereby enhancing grasp performance. To further plan the sections of the CCR other than the end-effector, the optimization objective $\Pi$ can be defined to minimize the distances $D_{(i, j)} (j \in \{1, 2, ..., n_{\text{dis}}\})$ between discrete points $j$ on the CR-$i$ and the corresponding discrete points on the target surface, while ensuring collision-free operation with the environment through a collision constraint $C_{\mathrm{coll}}$. This problem can be solved using the adaptive particle swarm optimization (PSO) algorithm, enabling complete conforming planning (Fig.~\ref{Fig_4}\textcolor{darkred}{(b)}) \cite{99}. In addition, inter-arm collision avoidance can be achieved either by removing collision-prone regions from the actuation space \cite{36}, or by discretizing each CR and enforcing a minimum inter-segment distance \cite{10}. However, when the object is deformable, the surface and contact state may evolve during interaction, limiting the applicability of these methods. Although such changes can be approximately modeled \cite{98}, accurately representing them and incorporating them into trajectory updating in a timely manner remain major challenges.

In assistance collaboration, where interaction between the main CR and the environment is typically considered, an RRT-based planner samples the CCR configuration space and, together with collision checking and spherical linear interpolation (SLERP), generates a feasible global trajectory for the main arm (Fig.~\ref{Fig_4}\textcolor{darkred}{(c)}) \cite{65}. However, under external interaction, task-space-level planning alone is often insufficient to satisfy coordination requirements in the chain space, especially under complex coupling conditions. This issue is more pronounced in parallel collaboration, where planning is typically centered on the moving platform \cite{73} and the coupling constraints are generally more complex.

Notably, existing motion planning methods for CCRs still largely adopt or extend traditional approaches used for individual CRs, such as PSO-based \cite{99} and RRT-based methods \cite{65}. Although effective for path generation in structured environments, these methods become difficult to scale efficiently as the number of CRs and coupling constraints increases, due to the growing complexity of search-space exploration, collision checking, and constraint handling. In addition, the adaptability of the above approaches to dynamic environments may remain limited, as the trajectories are generally predefined, which can in turn constrain their capability for autonomous task execution.

\begin{figure*} [t] 
	\centering
	\includegraphics[scale=0.5]{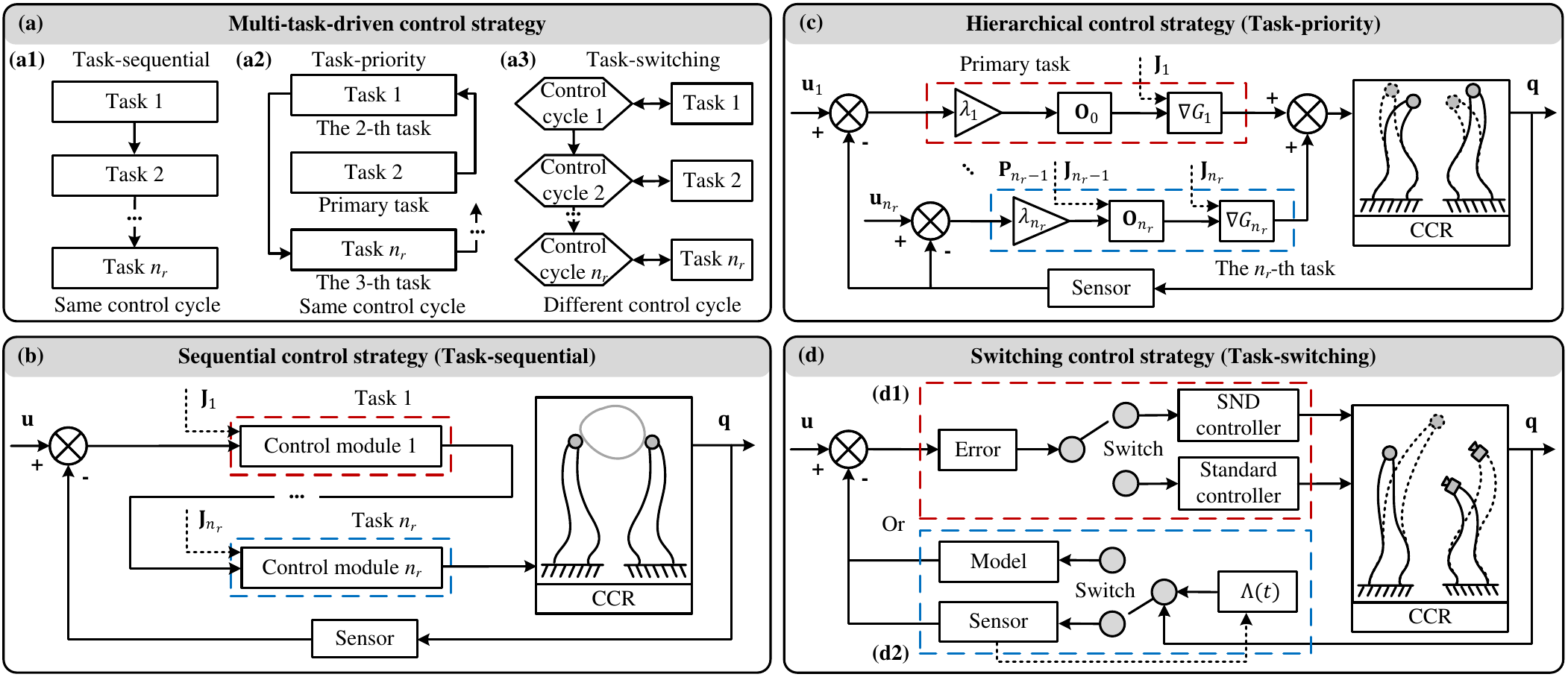} 
	\centering 
	\caption{Control of CCRs. (a) Multi-task-driven control strategy. (b) Sequential control strategy. (c) Hierarchical control strategy. (d) Switching control strategy.}
	\label{Fig_5} 
\end{figure*}

\subsubsection*{2) Online motion planning}

To enhance the adaptability of CCRs in dynamic environments, online motion planning can be adopted. To date, online motion planning has only been demonstrated in the separated collaboration mode, where real-time sensing is used to update CR states during each control cycle. According to the basis of online trajectory updating, existing methods can be categorized into goal-driven trajectory updating \cite{32,62} and reference-driven trajectory updating \cite{33,21,49} (Fig.~\ref{Fig_4}\textcolor{darkred}{(d)}).

In goal-driven trajectory updating, sensor feedback is continuously used to compute the pose deviation of each CR end-effector with respect to its corresponding goal point: $\Delta \boldsymbol{\delta}_i(t) = [\Delta \mathbf{P}_i(t), \Delta \mathbf{R}_i(t)]^\mathbf{T}$. Based on this deviation, the desired velocity is planned as \cite{6}
\begin{equation}
\hat{\boldsymbol{\delta}}_i(t) = f_{\delta} (\Delta \boldsymbol{\delta}_i(t), \hat{\boldsymbol{\delta}}_{(i,\mathrm{goal})}(t)),
\end{equation}
where $\hat{\boldsymbol{\delta}}_{(i,\mathrm{goal})}(t) = [\mathbf{v}_{(i,\mathrm{goal})}(t), \boldsymbol{\omega}_{(i,\mathrm{goal})}(t)]^\mathbf{T}$ denotes the linear and angular velocities of the goal point. By contrast, in reference-driven trajectory updating, the main CR follows a predefined offline trajectory or is operated manually, while the remaining CRs use sensor feedback to estimate the pose deviation \(\Delta \boldsymbol{\delta}_i(t)\) between the current pose of each CR and the corresponding reference point on the main CR in real time \cite{16}. However, existing approaches remain relatively limited, and their applicability to more strongly coupled collaboration modes still requires further investigation.

Overall, motion planning for CCRs is evolving from predominantly offline planning \cite{9} toward online planning \cite{6}, which improves environmental adaptability and supports more autonomous task execution. However, such adaptability still depends heavily on the accuracy, update rate, and robustness of the sensing mechanism. It is worth noting that motion planning studies on assistance and parallel collaboration remain relatively limited. One possible reason is that, under the coupling constraints imposed by the model, existing approaches often simplify the planning problem to trajectory generation of the master arm \cite{65} or the moving platform \cite{73}. However, such simplification is often insufficient to fully address the overall cooperative behavior of CCRs. Accordingly, online coordinated planning directly in the chain space remains a key challenge for CCRs, particularly in dynamic environments.

\subsection{Control for CCRs}
Once the target path/trajectory has been obtained, establishing an appropriate control strategy is essential to ensure precise task execution. Currently, most CCRs still employ open-loop control, as optimization in structural design can partially improve motion accuracy \cite{96}. A commonly used approach is the open-loop master–slave control strategy, in which a master device, captures the operator’s force signals and transmits them as motion commands to the slave CCR for remote execution of the corresponding actions \cite{30,31,38,39,90}. However, this approach relies on feedback derived from the operator’s perception, its performance is dependent on the operator’s experience. Moreover, in complex task environments, open-loop control may introduce  uncertainty.

To further enhance the performance of CCRs, control strategies should be designed based on sensing mechanisms \cite{106}. At present, most strategies focus on the separated collaborative mode and are typically model-based. Unlike single CRs, CCRs require synchronous control of multiple end-effectors to ensure coordinated task execution, which is usually achieved by combining the Jacobian matrices of the $n$ CRs into a unified Jacobian $\mathbf{J}$ \cite{37,32,8,9,10}. As a result, CCR control typically involves multiple tasks arising either within the same control cycle or across different control cycles, thereby posing greater challenges to control design and task coordination. On this basis, existing strategies can be broadly categorized as sequential, hierarchical, and switching strategy (Fig. \ref{Fig_5}\textcolor{darkred}{(a)} and Table \ref{Tab_4}).

For objectives within the same cycle, a straightforward approach is to adopt a sequential control strategy (Fig. \ref{Fig_5}\textcolor{darkred}{(a1)} and Fig. \ref{Fig_5}\textcolor{darkred}{(b)}), where different objectives are handled in order. For example, \cite{98} first minimizes the residual of end-effector pose and force–balance constraints before incorporating deformation energy minimization, whereas \cite{6} first enforces end-effector pose and closure constraints before minimizing equivalent joint torques. Although this strategy is straightforward to implement and can typically support a relatively high control frequency, earlier tasks may constrain the feasible space of subsequent tasks, thereby reducing their accuracy (Table \ref{Tab_4}). Moreover, errors may propagate along the task sequence, which can compromise robustness. To address this limitation more systematically, a hierarchical control strategy can be constructed by assigning importance levels to each task (Fig. \ref{Fig_5}\textcolor{darkred}{(a2)} and Fig. \ref{Fig_5}\textcolor{darkred}{(c)}) \cite{8}
\begin{equation}
	\mathbf{q} = \sum_{r=1}^{n_r} \lambda_{r} \,
	\mathbf{O}_{r-1} \, \nabla G_{r},
	\label{eq:control_law}
\end{equation}
where $\mathbf{q}$ denotes the system output, 
$\lambda_{r} (r=1, 2, ... , n_{r})$ is the weighting coefficient of task $r$, 
$\nabla G_{r}$ represents the control gradient of task $r$, 
and $\mathbf{O}_{r-1}$ is the null-space projection matrix of the first $(r-1)$ objectives. 
The projection matrix can be recursively constructed as
\begin{equation}
	\mathbf{O}_{r} = \mathbf{O}_{r-1} - 
	\left( \mathbf{J}_{r} \mathbf{O}_{r-1} \right)^{\dagger} 
	\left( \mathbf{J}_{r} \mathbf{O}_{r-1} \right), 
	\quad \mathbf{O}_{0} = \mathbf{I},
	\label{eq:projection}
\end{equation}
where $(\cdot)^{\dagger}$ denotes the Moore–Penrose pseudoinverse. 
By sequentially projecting each task into the null-space of higher-hierarchical objectives, 
this scheme enables the decoupling and coordination of multiple objectives, 
including end-effector relative position maintenance, inter-CR collision avoidance, trajectory tracking, 
and manipulability optimization \cite{9,10}. Compared with sequential control, this strategy is generally more effective in preserving the accuracy of high-priority tasks. However, the recursive computation of pseudoinverses and null-space projections increases the online computational burden and typically reduces the achievable control frequency. It is also more likely to cause numerical ill-conditioning under strong coupling or near singular configurations. Moreover, this strategy is mainly designed for multitask coordination within a single control cycle. When task priorities change across control cycles, a fixed priority hierarchy may no longer be sufficient. In such cases, switching control provides a more suitable alternative by introducing a threshold parameter to switch between control modes (Fig. \ref{Fig_5}\textcolor{darkred}{(a3)}). For instance, \cite{16} specifies that when the pose error exceeds threshold parameter, a normalized controller based on the normalized Jacobian $\mathbf{J}_{\mathrm{snd}}$ is employed to ensure faster convergence; otherwise, the system switches to a standard controller based on the standard Jacobian $\mathbf{J}_{s}$ to achieve higher accuracy (Fig. \ref{Fig_5}\textcolor{darkred}{(d1)}). Similarly, \cite{91} defines a perception state parameter $\Lambda(t)$, where $\Lambda(t)=1$ activates the closed-loop mode and $\Lambda(t)=0$ switches to the open-loop mode, thereby enhancing system robustness (Fig. \ref{Fig_5}\textcolor{darkred}{(d2)}). This strategy is well suited to changing operating conditions, although its performance is highly dependent on the switching criteria and transition smoothness. Inappropriate threshold settings may induce discontinuous inputs or chattering, thereby compromising control accuracy, numerical stability, and robustness.

\begin{table*}[t]
	\centering
	\caption{Comparison of control strategies for CCRs}
	\label{Tab_4}
	\setlength{\tabcolsep}{3pt}
	\renewcommand{\arraystretch}{1.05}
	\renewcommand{\tabularxcolumn}[1]{m{#1}}
	\begin{threeparttable}
		\begin{tabularx}{\textwidth}{c c c c c c >{\centering\arraybackslash}X >{\centering\arraybackslash}X c}
			\toprule
			\textbf{Mode} & 
			\textbf{Ref.} & 
			\textbf{Sensing mech.} & 
			\textbf{Open/Closed} & 
			\textbf{Control strategy} & 
			\textbf{Control frequency} & 
			$\boldsymbol{\varepsilon}_{\mathrm{pos}}$ & 
			$\boldsymbol{\varepsilon}_{\mathrm{ori}}$ & 
			$\boldsymbol{\varepsilon}_{\mathrm{for}}$ \\
			\midrule
			\multirow{12}{*}{SC}
			& {\cite{30}}  & VS & Open   & Master-slave & NR 
			& NR  & NR  & NR  \\
			& {\cite{31}}  & VS-FS-ES & Open   & Master-slave & NR  
			& NR  & NR  &NR  \\
			& {\cite{38}}  & VS-ES & Open   & Master-slave & NR  
			& 0.050\%$\sim$0.280\% 
			& 0.573\% & NR  \\
			& {[{\color{blue}128}]} & VS-FS & Open   & Master-slave & NR  
			& NR  & NR  & NR   \\
			& {\cite{90}}  & VS & Open   & Master-slave & NR  
			& 1.340\% & NR  & NR  \\
			& {\cite{98}}  & VS & Closed & Sequential & 10 Hz 
			& NR  
			& 0.430\%$\sim$0.462\% & NR  \\
			& {\cite{6}}   & Simulation & Closed & Sequential & 20 Hz 
			& 0.111\%$\sim$0.499\% 
			& 0.002\%$\sim$0.167\%
			& 0.025 N$\sim$0.340 N \\
			& {\cite{8}}   & ES & Closed & Hierarchical & NR  
			& LT: 0.118\%$\sim$0.355\% 
			& NR  & NR  \\
			& {\cite{9}}   & ES & Closed & Hierarchical & 1.6 Hz 
			& LT: 1.138\%$\sim$1.730\% & NR  & NR  \\
			& {\cite{10}}  & Simulation & Closed & Hierarchical & NR  
			& HT: 0.090\%$\sim$0.155\% & NR  & NR  \\
			& {\cite{16}}  & VS & Closed & Switching & NR  
			& NR  & NR  & NR  \\
			& {\cite{91}} & VS & Closed & Switching & NR  
			& 0.650 mm & 0.583\% & NR  \\
			\bottomrule
		\end{tabularx}
		\begin{tablenotes}
			\footnotesize
			\item SC: Separated Collaboration
			\item VS: Vision Sensor; FS: Force Sensor; ES: Electromagnetic Tracking Sensor
			\item $\boldsymbol{\varepsilon}_{\mathrm{pos}}$: Relative mean position error (Defined as the actual mean position error divided by the length of the corresponding CR)
			\item $\boldsymbol{\varepsilon}_{\mathrm{ori}}$: Relative mean orientation error (Defined as the actual mean angle error divided by $\pi$ rad)
			\item $\boldsymbol{\varepsilon}_{\mathrm{for}}$: Force error
            \item 
            HT: high-priority task; LT: low-priority task; NR: Not reported
		\end{tablenotes}
	\end{threeparttable}
\end{table*}

Control strategies for assistance and parallel collaboration remain relatively limited. Although existing strategies developed for separated collaboration, such as \cite{9, 10, 16}, may serve as useful references, they are generally difficult to extend directly to these two modes. One possible reason is that these two modes involve stronger inter-CR coupling \cite{73,77,78}, which reduces the motion independence of individual CRs and makes feasible control-input allocation more difficult. Moreover, under model inaccuracies, the controller may need to further incorporate sensing information such as connection-consistency errors or closed-chain constraint residuals as additional control tasks. This increases the complexity of unified-Jacobian-based multi-task computation and makes the associated pseudoinverse and null-space projection calculations more susceptible to numerical ill-conditioning, thereby imposing higher demands on the real-time capability, numerical stability, and robustness of the controller.

Overall, control strategies for CCRs increasingly emphasize multitask coordination. However, existing studies still focus mainly on relatively simple collaborative settings and remain limited in handling multitask coordination across control cycles, especially in assistance and parallel collaboration modes.

\subsection{Sensing mechanisms for CCRs} 
The key challenge in online motion planning and closed-loop control is accurately estimating and feeding back each CR’s pose in real time to enable error computation and actuation updates \cite{146}. 

\begin{figure} [t] 
	\centering
	\includegraphics[scale=0.5]{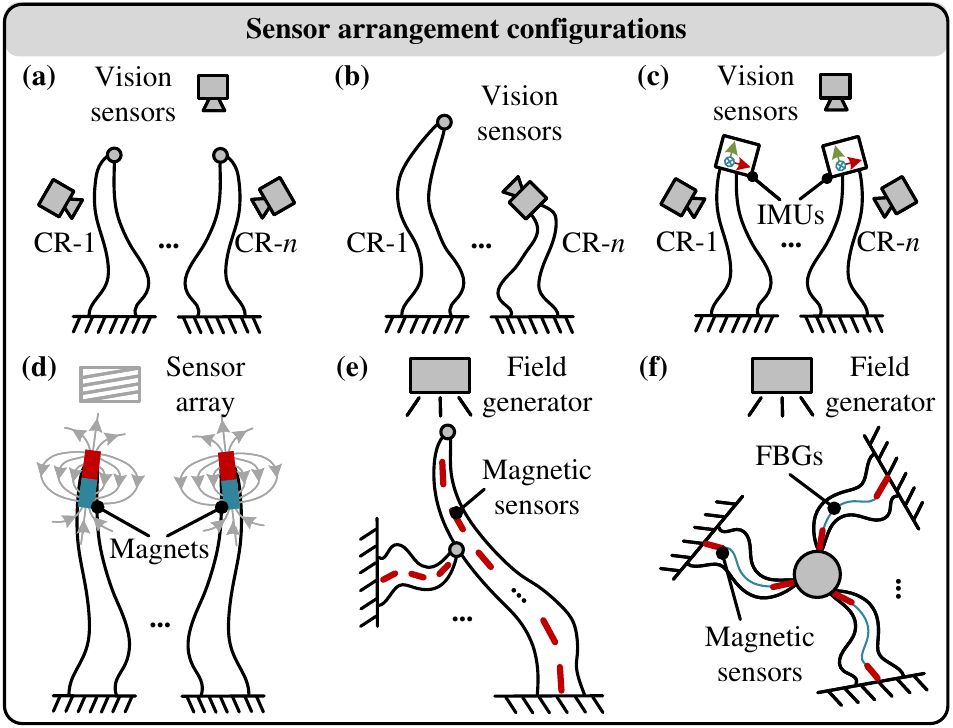} 
	\centering 
	\caption{Sensor arrangement configurations. (a) Eye-to-hand. (b) Eye-in-hand. (c) Vision–IMU. (d) External magnetic sensing. (e) Distributed magnetic sensing. (f) Magnetic–FBG sensing.}
	\label{Fig_6} 
\end{figure}

Vision-based sensing is the most widely used approach across the three collaboration modes, with sensor placement typically following two configurations: eye-to-hand (Fig. \ref{Fig_6}\textcolor{darkred}{(a)}) \cite{32,62} and eye-in-hand (Fig. \ref{Fig_6}\textcolor{darkred}{(b)}) \cite{33,21,49}. Pose estimation can be achieved either by detecting fiducial markers mounted on the CR surface \cite{6,62,44,91,92,137,139} or by applying deep learning algorithms to markerless CR images \cite{140,87}. When combined with geometric parameter identification, these methods can further enhance measurement accuracy \cite{73}. However, a major challenge remains the synchronized perception of multiple CRs, as the accuracy of visual sensing is highly sensitive to inter-CR occlusions, environmental obstacles, and poor lighting conditions.

To address these challenges, two representative strategies have been proposed. First, visual sensors can be combined with inertial measurement units (IMUs) mounted on each CR to estimate the end-effector pose (Fig. \ref{Fig_6}\textcolor{darkred}{(c)}), thereby improving robustness against external disturbances \cite{1}. Second, embedded sensors can be integrated with state estimation methods to achieve CCR pose perception \cite{210}. In the separated cooperation mode, \cite{44} embedded permanent magnets at each CR end-effector and employed an external magnetic sensor array (Fig. \ref{Fig_6}\textcolor{darkred}{(d)}). By fusing a PCC prior with a multi-dipole magnetic model in an extended Kalman filter, thereby enabling real-time tracking of multiple CR end-effectors. In the assistance mode, \cite{61} placed magnetic sensors discretely along the backbone (Fig. \ref{Fig_6}\textcolor{darkred}{(e)}). By combining a Cosserat-rod prior with forward extended Kalman–Bucy filtering over arc length and Rauch–Tung–Striebel smoothing, thereby enabling shape estimation under unknown end-effector loads. Furthermore, \cite{56} combined magnetic sensors at the base and end-effector with fiber Bragg grating (FBG) strain sensors along the backbone (Fig. \ref{Fig_6}\textcolor{darkred}{(f)}). By introducing a constant-strain Gaussian process prior on $SE(3)$ and performing maximum a posteriori estimation within a sparse factor-graph framework, coupling constraints were encoded as factors, thereby enhancing the real-time performance and robustness of sensing. The method can also be extended to parallel cooperation. Notably, further efforts are required to integrate these sensing mechanisms with online motion planning and control frameworks to fully realize their advantages. 

Overall, CCR sensing mechanisms are evolving from independent perception of individual CRs toward model-informed state estimation of collaborative coupling states. By integrating multi-source sensing with model priors, these approaches can improve the estimation of coupled deformation, connection states, and unknown loads, especially in assistance and parallel collaboration where inter-CR coupling is stronger. However, their implementation still relies heavily on external sensing devices or complex embedded sensor layouts, complicating sensor integration and calibration and thereby potentially limiting deployment in complex environments.

\section {CHALLENGES AND OPPORTUNITIES}
\label{section5}
Despite substantial progress in CCRs over the past two decades, many challenges remain unresolved. In this section, research opportunities in key technologies are first discussed, followed by system integration challenges for real-world deployment. Finally, potential applications of CCRs are outlined.

\subsection{{Research opportunities in key technologies}}
Future opportunities corresponding to the key technical challenges identified in Sections \ref{section_xz}–\ref{section4} are discussed below.

\subsubsection{Autonomous multi-CR task allocation}
To address the lack of systematic automated mechanisms for multi-CR task allocation (Section \ref{section_xz}\textcolor{darkred}{-A}), future research may formulate multi-CR task allocation as a constrained optimization problem that accounts for CR compliance and inter-CR coupling, and combine multi-objective optimization and reinforcement learning to achieve autonomous allocation. Statistically distinct planning \cite{204} can be used to generate multiple feasible and differentiated candidate allocation plans, thereby supporting online reallocation when predefined schemes become infeasible and improving the robustness of task execution.

\subsubsection{Generative-model-driven structural design}
To reduce the dependence of current design processes on designer expertise (Section \ref{section2}\textcolor{darkred}{-E}), future research may introduce generative models to generate candidate CCR configurations under application-specific and structural constraints. In particular, deep-level requirement mining in design theory \cite{174} may provide design inputs derived from application requirements, while a knowledge graph could be constructed to systematically summarize reusable design experience reported in CCR publications \cite{171,172}, thereby supporting knowledge-guided generative structural design.

\subsubsection{Physics-informed data-driven modeling}
To address the trade-off between the physical fidelity of inter-CR interaction modeling and computational efficiency (Section \ref{section3}\textcolor{darkred}{-D}), physics-informed neural networks (PINNs) offer a direction by embedding physical constraints into the loss function, thereby improving physical consistency and reducing reliance on large experimental datasets \cite{149}. While PINNs have demonstrated effectiveness in single-CR modeling \cite{148}, their applicability to strongly coupled CCRs remains underexplored.

\subsubsection{Reinforcement-learning–based planning}
To address the challenge of online coordinated planning in the chain space in dynamic environments (Section \ref{section4}\textcolor{darkred}{-A}), reinforcement learning \cite{144,175,181,182} may provide a promising approach. By interacting with the environment, such methods can learn adaptive planning policies that balance cooperation and obstacle avoidance without heavily relying on precise models or extensive preprocessing. At the algorithmic level, integrating imitation learning \cite{143} may help reduce sample complexity, while incorporating game-theoretic methods \cite{145} may improve the robustness of multi-CR cooperative decision-making.

\subsubsection{Task-decoupled and adaptive control}
To improve multitask coordination across control cycles (Section \ref{section4}\textcolor{darkred}{-B}), natural task-space decoupling \cite{183} may be introduced within each control cycle to reduce reliance on strictly manually assigned task priorities while enhancing control stability. Building on this, a multi-criteria adaptive threshold-switching mechanism may help reduce reliance on empirically fixed thresholds, thereby enabling the dynamic reconfiguration of task sets and control modes across control cycles.

\subsubsection{Intelligent-material-based self-sensing}
To address the sensor integration and calibration challenges (Section \ref{section4}\textcolor{darkred}{-C}), future research may explore intelligent materials with piezoresistive, capacitive, or optical responses \cite{173,180,194,207}, which may be embedded into CR bodies to form distributed sensing units. By establishing mappings between material responses and the deformation and interaction states of CRs, accurate estimation of system states may be achieved. This could reduce reliance on external sensing systems, improve system integration and environmental adaptability, and provide reliable state feedback for online motion planning and control.

Overall, generative-model-driven structural design provides candidate CCR configurations and collaboration modes, while physics-informed data-driven modeling further establishes the corresponding models for state prediction and performance evaluation. On this basis, autonomous multi-CR task allocation selects suitable CR subsets and matches them to tasks, reinforcement-learning-based planning generates coordinated action sequences for the assigned CR subsets, and task-decoupled adaptive control translates the planned actions into actual execution by individual CRs. During execution, intelligent-material-based self-sensing continuously acquires system-state information and feeds it back to autonomous multi-CR task allocation, reinforcement-learning-based planning, and task-decoupled adaptive control. Together, these technologies are expected to enhance the overall performance of closed-loop task execution in CCRs.

\subsection{{System integration challenges for real-world deployment}}
Although promising potential for medical and industrial applications has been demonstrated, CCRs have largely remained at the research and laboratory-prototype stages \cite{89,98,2,6}. With continued advances in the aforementioned key technologies, reliable integration across the functional, execution–control, and decision levels of the closed-loop task-execution architecture introduced in Section \ref{section_xz} is required for real-world deployment. The main integration challenges and possible approaches to addressing them are discussed below.

\subsubsection{{Runtime task supervision and recovery}}
During real-world deployment, prompt assessment of whether the current task allocation remains feasible may be hindered by incomplete or inconsistent feedback provided by sensing mechanisms at the functional level and task-execution feedback from the execution-control level. Runtime monitoring and recovery methods \cite{199,200,201} could therefore be adapted to continuously assess task feasibility and safety risks, with task reallocation triggered when the current allocation becomes infeasible.

\subsubsection{{Latency-aware temporal alignment}}
During real-world deployment, delays in the transmission and processing of multisource sensory information along the execution-control chain may cause state feedback to lag behind actual task execution, thereby degrading multi-CR coordination performance \cite{44,61,56}. Latency-aware temporal alignment mechanisms \cite{205} should therefore be developed to temporally align state feedback with coupled planning and coordinated control throughout closed-loop task execution.

\subsubsection{{Physical integration and multi-frame calibration}}
During real-world deployment, CR bodies, sensing mechanisms, cables, and interfaces must be integrated within constrained spaces, while inadequate packaging and cable routing may cause mechanical interference. Calibration errors among the base, end-effector, and task frames may also reduce the accuracy of multi-CR coordination \cite{203}. Compact integration schemes and unified multi-frame calibration procedures should therefore be developed to improve deployment reliability.

Together, these directions address task-level adaptation, temporal consistency, and physical reliability, respectively, thereby providing the system-level basis for real-world CCR deployment. Beyond these integration requirements, although current CCR studies involve only a small number of CRs \cite{81,88} and have not yet formed large-scale multi-agent systems in the conventional sense, the scalable deployment of CCRs may exhibit multi-agent-system characteristics as the number of CRs and coordination complexity increase \cite{190, 191}.

\subsection{Potential applications}
As the system-integration challenges for real-world deployment are progressively addressed, CCRs are expected to be deployed across a broader range of domains requiring compliant cooperative manipulation and confined-space operation. Representative potential applications are discussed below.
\subsubsection{Rehabilitation} Experiments have shown that interactions such as hugging with rigid collaborative robots sheathed in soft materials \cite{147} can effectively reduce patient stress \cite{150}. Introducing CCRs with higher compliance could further broaden the range of safe interactions between humans and robots and thereby improve the effectiveness of rehabilitation training, showing considerable promise.

\subsubsection{Emergency response} Disaster-relief environments are often narrow with tortuous paths, and conventional articulated robots, constrained by rigid links and limited joint ranges, struggle to access target areas \cite{176,178}. Leveraging the compliance and collaborative capabilities of CCRs could enable environmental probing, payload delivery, in-cavity disassembly, and rescue operations in such extreme settings, thereby increasing the likelihood of successful rescue.

\subsubsection{Agriculture} In fruit and vegetable harvesting, the high compliance of CCRs allows them to reach into gaps between leaves and branches \cite{177,179}, while maintaining conformity to delicate plants.  With onboard shape and force sensing, real-time perception and feedback control could be implemented, with the potential to reduce damage risk to crops and thereby improve harvesting efficiency.

\section {CONCLUSION}
\label{section6}

CCRs offer advantages over single CRs, including enhanced task adaptability, flexibility, and load capacity, enabling a wider range of applications. This survey constructs a closed-loop task-execution architecture for CCRs from an automation-system perspective and classifies three representative collaboration modes. Research progress is summarized and analyzed in terms of key technologies, including multi-CR task allocation, structural design, modeling, motion planning methods, control strategies, and sensing mechanisms.

Future CCR research should advance both key technologies and deployment-oriented capabilities. In addition to autonomous multi-CR task allocation, generative-model-driven structural design, physics-informed data-driven modeling, RL-based planning, task-decoupled and adaptive control, and intelligent-material-based self-sensing, greater attention should be paid to runtime task supervision and recovery, latency-aware temporal alignment, and physical integration and multi-frame calibration. These efforts are expected to enhance CCR performance and promote the transition from laboratory-scale prototypes to real-world deployment, thereby supporting automation applications in surgery, rehabilitation, industry, emergency response, and agriculture.

\bibliographystyle{IEEEtran}
\bibliography{citationlist}

\vspace{-40pt}

\begin{IEEEbiography}[{\includegraphics[width=1in,height=1.25in,clip,keepaspectratio]{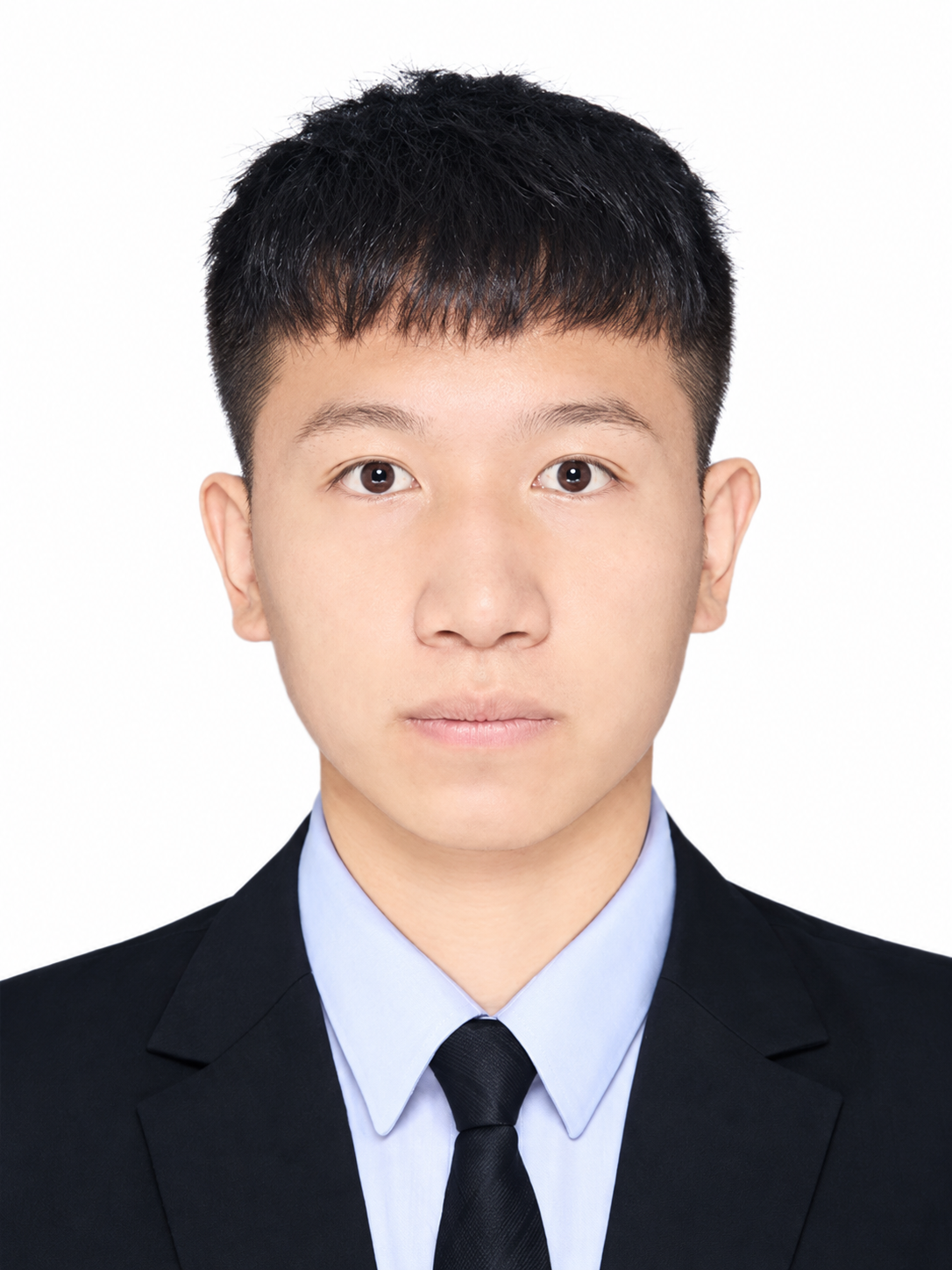}}]{Xinyu Li}
(Graduate Student Member) received the M.E. degree in mechanical engineering from Hebei University of Technology, Tianjin, China, in 2023. He is currently working toward the Ph.D. degree in mechanical engineering in the College of Mechanical and Vehicle Engineering, Chongqing University, Chongqing, China. 

His current research interests include continuum robotics and innovative design theory.
\end{IEEEbiography}

\vspace{-25pt}

\begin{IEEEbiography}[{\includegraphics[width=1in,height=1.25in,clip,keepaspectratio]{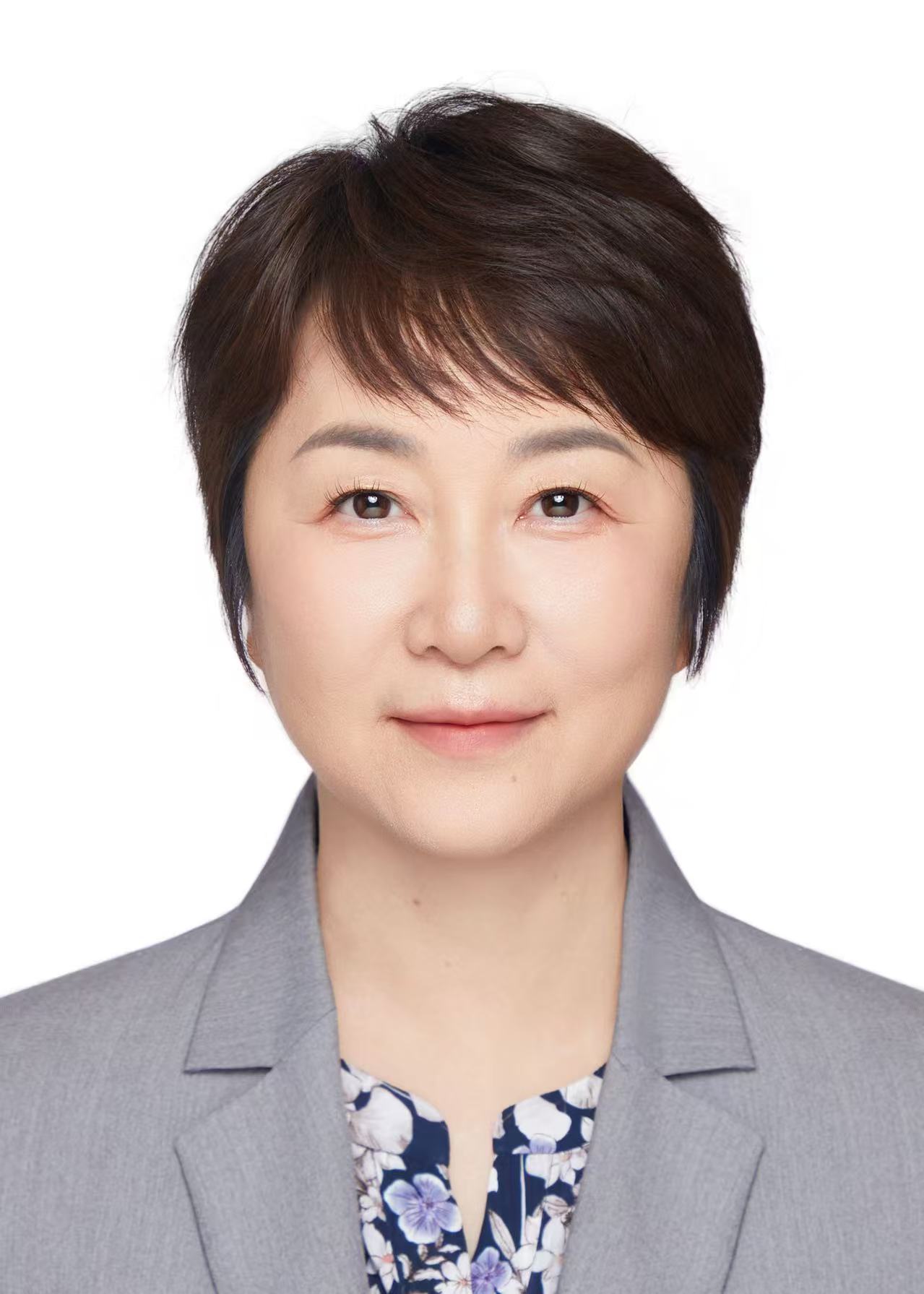}}]{Qian Tang}
	received the B.S., M.S., and Ph.D. degrees in mechanical engineering from Chongqing University, Chongqing, China, in 1991, 1994, and 1997, respectively. She is currently a Professor with the College of Mechanical and Vehicle Engineering, Chongqing University. 
    
    Her current research interests include robotics, mechanical transmissions theory and technology, intelligent manufacturing, and additive manufacturing.
\end{IEEEbiography}

\vspace{-25pt}

\begin{IEEEbiography}[{\includegraphics[width=1in,height=1.25in,clip,keepaspectratio]{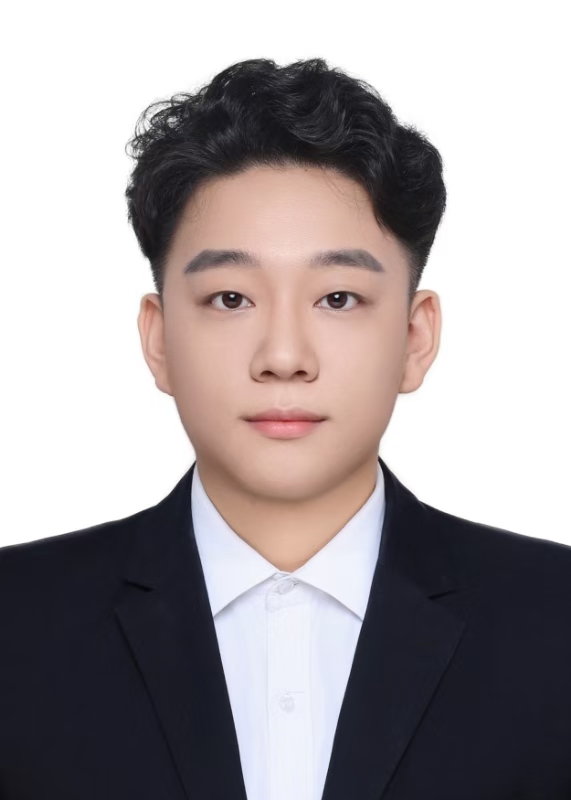}}]{Guoxin Yin}
received the B.S. degree in robot engineering from the Chongqing University of Technology, Chongqing, China, in 2024. He is currently working toward the M.S. degree in robot engineering with the Chongqing University, Chongqing, China. 

His research interests include design, modeling, and control of continuum robots.
 \end{IEEEbiography}

\vspace{-20pt}

\begin{IEEEbiography}[{\includegraphics[width=1in,height=1.25in,clip,keepaspectratio]{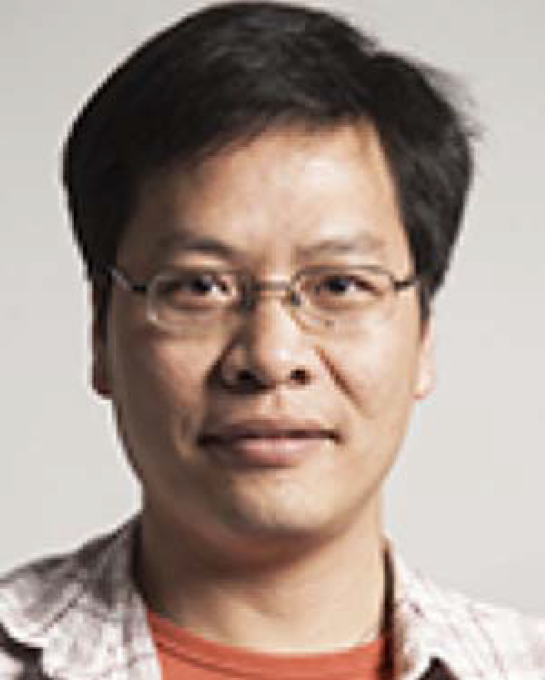}}]{Gang Zheng}
	(Senior Member) received the B.E. and M.E. degrees in communication and systems from Wuhan University, Wuhan, China, in 2001 and 2004, respectively, and the Ph.D. degree in automatic control from ENSEA, Cergy-Pontoise, France, in 2006. Since 2007, he has held Postdoctoral positions with INRIA Grenoble, Montbonnot-Saint-Martin, France, with the Laboratoire Jean Kuntzmann, Paris, France, and with ENSEA, Cergy, France. He joined INRIA Lille as a Permanent Researcher from 2009. 
    
    His research interests include control and observation of nonlinear systems, and its applications to rigid and soft robotics.  
\end{IEEEbiography}

\vspace{-20pt}

\begin{IEEEbiography}[{\includegraphics[width=1in,height=1.25in,clip,keepaspectratio]{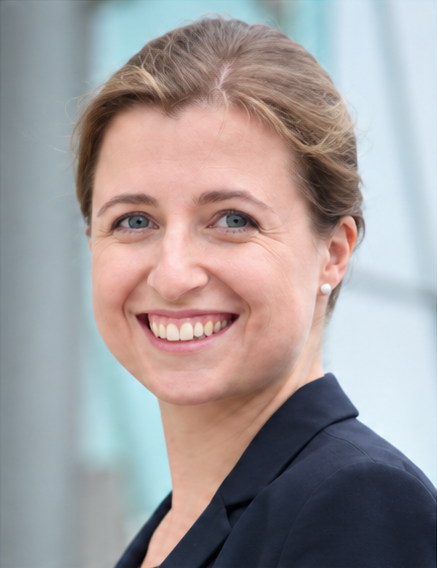}}]{Jessica Burgner-Kahrs}
	(Senior Member) received the diploma and the Ph.D. degree in computer science from the Karlsruhe Institute of Technology (KIT), Karlsruhe, Germany, in 2006 and 2010, respectively.

She is a Professor with the Departments of Mathematical and Computational Sciences, Computer Science, and Mechanical and Industrial Engineering, the founding Director of The Continuum Robotics Laboratory, and Associate Director of the Robotics Institute, University of Toronto, Toronto, ON, Canada. From 2013 to 2019, she was with Leibniz University Hannover, Hanover, Germany, and from 2010 to 2012, with Vanderbilt University, Nashville, TN, USA. Her fundamental robotics research is driven by applications in minimally invasive surgery and maintenance, repair, and operations. Her research interests include continuum robotics and in particular on their design, modeling, planning, and control, as well as human--robot interaction.  
\end{IEEEbiography}

\vfill
\newpage

\begin{IEEEbiography}[{\includegraphics[width=1in,height=1.25in,clip,keepaspectratio]{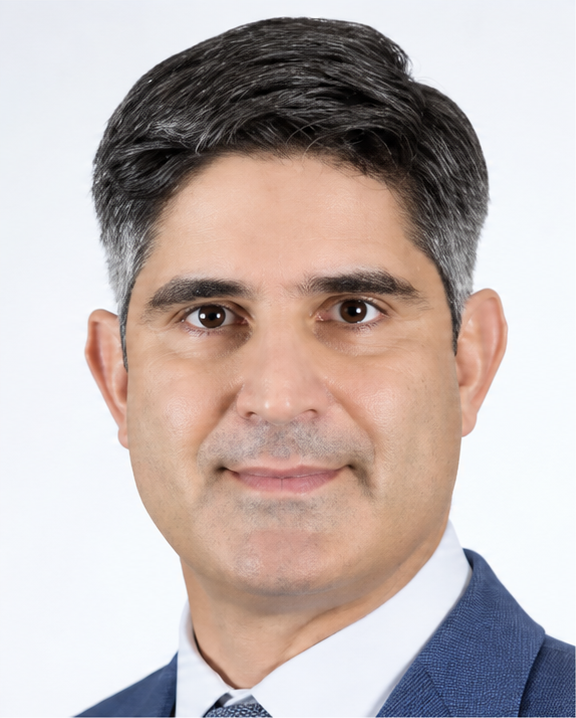}}]{Cesare Stefanini}
	(Member) received the M.Sc. (Hons.) degree in mechanical engineering and the Ph.D. (Hons.) degree in microengineering from Scuola Superiore Sant'Anna (SSSA), Pisa, Italy, in 1997 and 2002, respectively.

He is currently a Professor with the Department of Robotics, Mohamed bin Zayed University of Artificial Intelligence, Abu Dhabi, United Arab Emirates. He is the author or coauthor of more than 370 articles in refereed international journals and in international conference proceedings. He is the inventor of more than 15 international patents, nine of which have been industrially exploited by international companies. His research interests include underwater robotics, bioinspired systems, biomechatronics, and micromechatronics for medical and industrial applications.
\end{IEEEbiography}

\vspace{-20pt}

\begin{IEEEbiography}[{\includegraphics[width=1in,height=1.25in,clip,keepaspectratio]{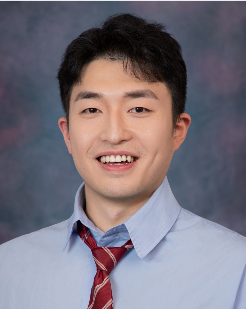}}]{Ke Wu}
	(Member) received the B.S. degree in mechanical engineering from Chongqing University, China, in 2016, and dual M.E. degrees in mechanical engineering from Chongqing University and University College Dublin, Ireland, in 2019. He completed his Ph.D. in robotics and applied mathematics at INRIA Lille, France, in 2023. He joined Mohamed bin Zayed University of Artificial Intelligence (MBZUAI), United Arab Emirates, in 2024, where he is currently a Visiting Assistant Professor of Robotics. 
    
    His research focuses on robotics and continuum robotics, with an emphasis on physics-based simulation, robot learning, and embodied AI. 
\end{IEEEbiography}

\vspace{11pt}

\vfill

\end{document}